\documentclass[sigconf]{acmart}

\AtBeginDocument{%
  }

\usepackage{multirow}
\usepackage{multicol}
\usepackage{pifont}

\usepackage{color, colortbl}
\usepackage{hyperref}

\copyrightyear{2024}
\acmYear{2024}
\setcopyright{acmlicensed}
\acmConference[MM '24] {Proceedings of the 32nd ACM International Conference on Multimedia}{October 28--November 1, 2024}{Melbourne, VIC, Australia.}
\acmBooktitle{Proceedings of the 32nd ACM International Conference on Multimedia (MM '24), October 28--November 1, 2024, Melbourne, VIC, Australia}
\acmISBN{979-8-4007-0686-8/24/10}
\acmDOI{10.1145/3664647.3681537}
\begin{document}

\title{Probabilistic Vision-Language Representation for Weakly Supervised Temporal Action Localization}


\author{Geuntaek Lim}
\affiliation{%
  \institution{Sejong University}
  \city{Seoul}
  \country{Republic of Korea}}
\email{gtlim@rcv.sejong.ac.kr}

\author{Hyunwoo Kim}
\affiliation{%
  \institution{Sejong University}
  \city{Seoul}
  \country{Republic of Korea}}
\email{hwkim@rcv.sejong.ac.kr}

\author{Joonsoo Kim}
\affiliation{%
  \institution{Electronics and Telecommunications Research Institute}
  \city{Daejeon}
  \country{Republic of Korea}}
\email{joonsookim@etri.re.kr}

\author{Yukyung Choi}
\authornote{Yukyung Choi is the corresponding author.}
\affiliation{%
  \institution{Sejong University}
  \city{Seoul}
  \country{Republic of Korea}}
\email{ykchoi@rcv.sejong.ac.kr}

\renewcommand{\shortauthors}{Geuntaek Lim, Hyunwoo Kim, Joonsoo Kim, \& Yukyung Choi}

\begin{abstract} Weakly supervised temporal action localization (WTAL) aims to detect action instances in untrimmed videos using only video-level annotations. Since many existing works optimize WTAL models based on action classification labels, they encounter the task discrepancy problem (\textit{i.e.}, localization-by-classification). To tackle this issue, recent studies have attempted to utilize action category names as auxiliary semantic knowledge through vision-language pre-training (VLP). However, there are still areas where existing research falls short. Previous approaches primarily focused on leveraging textual information from language models but overlooked the alignment of dynamic human action and VLP knowledge in a joint space. Furthermore, the deterministic representation employed in previous studies struggles to capture fine-grained human motions. To address these problems, we propose a novel framework that aligns human action knowledge and VLP knowledge in a probabilistic embedding space. Moreover, we propose intra- and inter-distribution contrastive learning to enhance the probabilistic embedding space based on statistical similarities. Extensive experiments and ablation studies reveal that our method significantly outperforms all previous state-of-the-art methods. Code is available at \href{https://github.com/sejong-rcv/PVLR} {\color{magenta}{https://github.com/sejong-rcv/PVLR}}.
\end{abstract}



\maketitle

\section{Introduction}

    The development of multimedia services such as YouTube and Netflix, has sparked growing interest in the field of computer vision for analyzing long-form videos. Temporal Action Localization (TAL) refers to the problem of precisely determining the time intervals in a lengthy, untrimmed video when human activities occur, which is fundamentally significant for video understanding~\cite{jo2023vvs, luo2022clip4clip, ma2022x}. 

    \begin{figure}[t]
        \centering

        \includegraphics[width=0.95\linewidth]{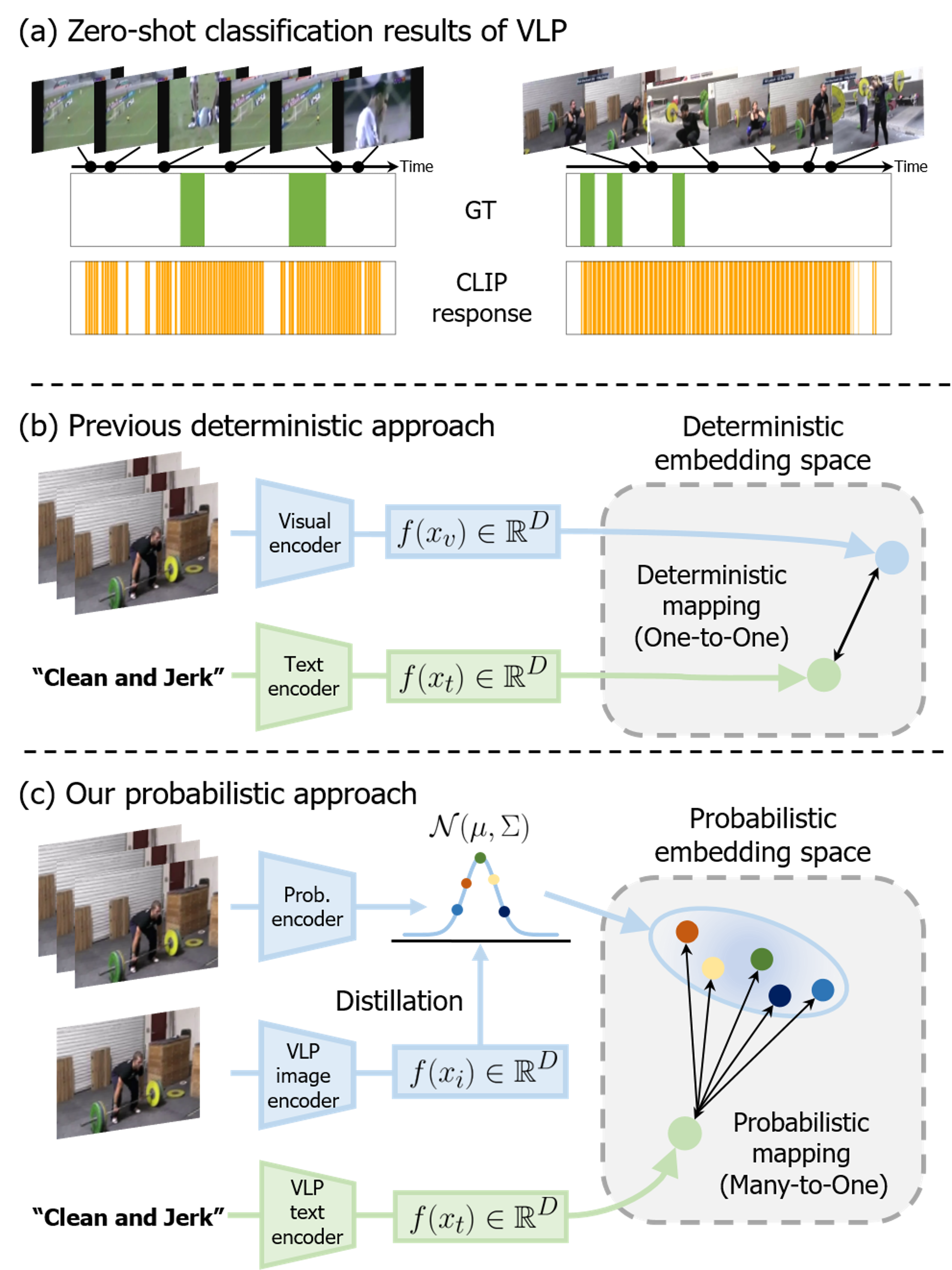}
        \captionsetup{justification=justified,singlelinecheck=false}
        \caption{(a) CLIP's deterministic pre-training with image-text pairs fails to equip it with the necessary understanding of fine-grained human motion variations. (b) Earlier studies have primarily focused on the direct mapping between language models and visual input based on deterministic representation. (c) The proposed framework utilizes probabilistic embedding and aligns VLP knowledge.
        \vspace{-8mm}}
        \label{fig:fig1}
        \Description[Motivation of our work]{The depicted figure encapsulates the motivation for this research. The figure at the top illustrates the inadequacy of traditional Vision-Language Pre-training methods for human action comprehension. The figure in the middle offers a concise introduction to frameworks from previous research that leverage text information. The bottommost figure represents the framework proposed by us.}
    \end{figure}

    Fully supervised TAL~\cite{cheng2022tallformer, zhang2022actionformer, tirupattur2021modeling, chao2018rethinking, lin2019bmn, liu2019novel,zhang2022temporal} handles this task using rich frame-level annotations. Despite its success, training a TAL model with dense frame-level annotations poses challenges due to the high annotation costs and limited generality. To address these challenges, weakly supervised temporal action localization (WTAL)~\cite{nguyen2018weakly, lee2020background, li2023boosting, shen2020weakly, kim2022background, he2022asm, hu2023learning}, which only requires video-level categorical labels, has received significant attention. In scenarios with only video-level category supervision, existing WTAL methods address the localization problem by selecting discriminative snippets\footnote{In our field, a snippet refers to a set of consecutive frames composed of 16 frames.} primarily for video-level classification. However, these classification-based approaches suffer from the task discrepancy problem inherent in a localization-by-classification framework. To tackle WTAL through a localization-by-localization framework, numerous studies~\cite{li2022exploring, huang2022weakly, rizve2023pivotal, zhou2023improving} has focused on snippet-level pseudo-label methods. Yet, these pseudo labels, constrained by video-level annotations, inherently carry noisy proposals and fail to achieve the desired accuracy.
    
    Recently, some approaches~\cite{li2023boosting, ju2023distilling} leveraging action category text information have been proposed to address these problems by incorporating powerful semantic knowledge without incurring annotation costs. These approaches establish additional learning cues by exploiting category text embedding vectors instead of merely utilizing category information as one-hot vectors. While significant improvements were made through additional semantic information, some factors were overlooked in earlier research. \textbf{First,} Li \textit{et al.}~\cite{li2023boosting} adopted a language model (\textit{e.g.}, GloVe~\cite{pennington2014glove}) pre-trained solely on the text modality, resulting in inadequate initialization for alignment with the human action pre-trained visual features. \textbf{Second,} Chen \textit{et al.}~\cite{ju2023distilling} proposed an alternative optimization strategy to introduce an effective distillation framework. However, this alternative optimization scheme requires manually identifying the optimal settings according to the dataset. \textbf{Third,} and most importantly, we observe that the utilization of deterministic representation in previous studies for incorporating text information is not optimal for human action understanding. 
    
    To confirm this, we conducted an analysis of zero-shot classification with CLIP~\cite{CLIP}, a prominent study in the domain of VLP. As shown in Figure~\ref{fig:fig1}(a), we compared the similarity response between the text prompt (\textit{i.e.}, 'a frame of [CLS]') representation and the corresponding frame visual representation. The analysis reveals a high level of activation even when actual human actions do not occur, as long as there is visual relevance to the action text category. This is because CLIP is pre-trained to consider only one-to-one matching between a single image and its caption. The previous research depicted in Figure \ref{fig:fig1}(b) cannot address this issue as it solely relies on deterministic representation via one-to-one matching, making it challenging to capture fine-grained human motion. Furthermore, the lack of consideration for direct alignment with pre-trained human action knowledge results in insufficient modeling of temporal dynamics.

    To overcome this issue, we introduce a novel framework, \textbf{PVLR}, \textbf{P}robabilistic \textbf{V}ision \textbf{L}anguage \textbf{R}epresentation for Weakly supervised Temporal Action Localization, which integrates VLP knowledge and human action knowledge within the probabilistic embedding space, as shown in  Figure~\ref{fig:fig1}(c). To begin with, pre-trained human action knowledge, such as Kinetics~\cite{i3d}, is utilized to initialize a probabilistic embedding space. In this step, probabilistic adapters are introduced to estimate parameters for the snippet-level probability distributions. Subsequently, we transfer the large-scale VLP knowledge to the estimated probability distribution to learn a joint probabilistic embedding space. To capture the temporal dynamics of action, we obtain samples from the estimated probability distribution to offer diverse perspectives, perform many-to-one matching, then measure their similarity with category text embeddings via Monte-Carlo estimation.

    Furthermore, to learn a distinctive embedding space, we propose a distribution contrastive learning scheme to capture the statistical similarities between distributions. We enhance intra-class compactness by learning the similarity of content (action or background) within videos and maximize inter-class separability by leveraging action category information across videos. To enhance the intra-class compactness, we draw inspiration from snippet mining in prior work~\cite{zhang2021cola} to differentiate similar snippet distributions among related content. For inter-class separability, we build a video-level probabilistic distribution based on a Gaussian mixture model (GMM) and make the mixture distribution separable between different action classes. To the best of our knowledge, this is the first attempt to investigate multi-modal probabilistic representations for weakly supervised temporal action localization. Our main contributions in this work are summarized as follows:
    
        \begin{enumerate}
        \item We introduce a novel framework that aligns VLP knowledge and action knowledge within a probabilistic space to fully consider temporal dynamics for fine-grained motion modeling.
        \item We propose an intra- and inter-distribution contrastive strategy based on statistical distance to construct a distinctive probabilistic embedding space.  
        \item We conduct extensive experiments and ablation studies to demonstrate the significance of the probabilistic embedding and the proposed method, showing superior performance on two public benchmarks, THUMOS14 and ActivityNet v1.3. 
        \end{enumerate}

\section{RELATED WORK}
    
    \subsection{Weakly Supervised Temporal Action Localization} 
        Weakly supervised temporal action localization (WTAL) is proposed to alleviate the laborious annotation procedure for Temporal Action Localization, training with only video-level labels.
        In the early stages of research, Multiple Instance Learning (MIL)-based approaches~\cite{nguyen2018weakly, lee2020background, lee2021weakly, islam2021hybrid, shou2018autoloc, paul2018w, hong2021cross} were proposed, treating a video as a bag of multiple action and background instances. Zhang \textit{et al.}~\cite{zhang2021cola} introduced a snippet contrast loss, refining the representation of ambiguous instances in the feature space through snippet mining and contrastive learning. Subsequently, several approaches~\cite{huang2022weakly, zhou2023improving, rizve2023pivotal, li2022exploring} have generated snippet-level pseudo labels to explicitly guide the model as a localization-by-localization framework. However, pseudo labels generated based on video-level supervision were often inaccurate and noisy, making it challenging to achieve the desired performance.
            
        Recently, approaches utilizing the semantic information of action category names have emerged to address the fundamental absence of temporal annotation in WTAL~\cite{ju2023distilling, li2023boosting}. Li \textit{et al.}~\cite{li2023boosting} designed a novel framework with a discriminative objective to enlarge inter-class differences and a generative objective to enhance intra-class integrity via text information. Chen \textit{et al.}~\cite{ju2023distilling} proposed a novel distillation and collaboration framework with complementary Classification Based Pre-training (CBP) and Vision-Language Pre-training (VLP) branches. While these works distinguish themselves with promising performances without additional annotation costs, there is still potential for further development. In our framework, we integrate VLP knowledge and human action knowledge within a probabilistic space previously unexplored in existing literature, enabling a diverse understanding of human action. 
    
    \subsection{Vision Language Pre-training} 
        Vision language pre-training (VLP) learns a joint representation through large-scale image-text pair datasets with consistent contextual information. A representative work is CLIP~\cite{CLIP}, which maps image-text pairs that contain consistent contextual information into the visual and textual encoders separately, facilitating the learning of a joint embedding space through aligned representations. CLIP has shown great success in many image understanding tasks, including image classification~\cite{conde2021clip, novack2023chils}, semantic segmentation~\cite{rao2022denseclip, kwon2023probabilistic}, image generation~\cite{saharia2022photorealistic, crowson2022vqgan}, and visual question answering~\cite{parelli2023clip}. Building upon the success of CLIP in the image domain, some research efforts~\cite{luo2022clip4clip, ma2022x, xue2022clip, liu2023revisiting} have emerged that aim to leverage CLIP's vision-language representation in the video domain. Our work also contributes to the research aimed at extending VLP knowledge into the realm of untrimmed video and human action understanding. \vspace{-5mm}
     
    \subsection{Probabilistic Representation} 
        Probabilistic representation has shown a strong potential to embed asymmetric relations, quantify uncertainty, add robustness, and more. The main idea of probabilistic embedding is to map inputs to probability distributions in the embedding space. To achieve this, the desired distributions are estimated by a deep neural network and optimized to maximize their likelihood. PCME~\cite{chun2021probabilistic} models many-to-one relationships in the joint embedding space with uncertainty estimation and introduces a soft cross-modal contrastive loss. ProViCo~\cite{park2022probabilistic} proposed self-supervised video representation learning that bridges contrastive learning with probabilistic embedding with a Gaussian mixture model. ProbVLM~\cite{upadhyay2023probvlm} utilizes a probabilistic adapter that estimates probability distributions for embeddings of a vision-language pre-trained model through inter- and intra-modal alignment in a post-hoc manner, without requiring extensive datasets or intensive computation. The objective of this study is to transfer the knowledge of a pre-trained vision-language model into the probabilistic embedding space, with an explicit aim of enhancing human action understanding.

\section{METHOD}
    \begin{figure*}[ht!]
        \centering
        \includegraphics[width=\linewidth]{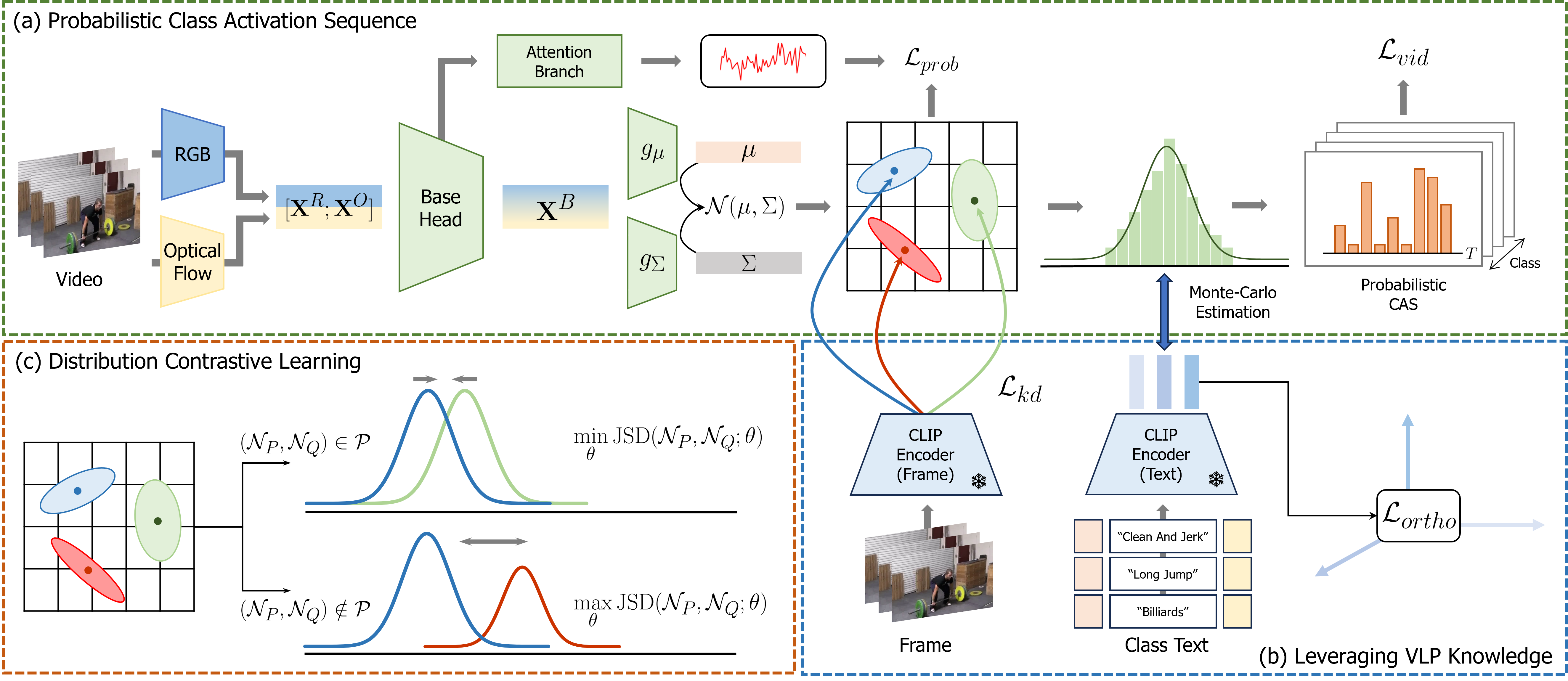}
        \captionsetup{justification=justified,singlelinecheck=false}
        \caption{Overview of the proposed PVLR. (a) \textbf{Probabilistic Class Activation Sequence}: For the probabilistic embedding, probabilistic adapters are augmented to facilitate the estimation of probabilistic distributions for individual snippets. (b) \textbf{Leveraging VLP knowledge}: We estimate probabilistic distributions and guide the model with semantic textual information corresponding to action categories. (c) \textbf{Distribution Contrastive Learning}: By training statistical similarities from probabilistic distribution, we aim to build distinctive embedding space.
        \vspace{0mm}} 
        \label{fig:fig2}
        \Description[Overall framework of our work]{The topmost figure describes the section detailing our proposed Probabilistic Class Activation Sequence. The bottom-left figure describes the section detailing our proposed Distribution Contrastive Learning. The bottom-right figure illustrates how Vision-Language Pre-training is utilized in our proposed framework.}
    \end{figure*}
    
    \subsection{Base Approach} \label{sec3a}
        \subsubsection{Base Head} \label{sec3aa}
            Our work introduces the probabilistic vision-language representation into a basic MIL approach. To begin, we detail the fundamental base head of the MIL approach in this section. We split each video into multi-frame, non-overlapping snippets and sample a fixed number $T$ of snippets to handle variations in video lengths. After sampling, it is common practice to use a pre-trained snippet feature extractor~\cite{i3d} for RGB $\mathbf{X}^{R}=\left\{\mathbf{x}^{r}_{t} \right\}_{t=1}^{T}$ and optical flow $\mathbf{X}^{O}=\left\{\mathbf{x}^{o}_{t} \right\}_{t=1}^{T}$ representation. Afterwards, we concatenate features from each modality $[\mathbf{X}^{R};\mathbf{X}^{O}] \in \mathbb{R}^{T\times 2D}$ and feed them into the base head $f_{base}$, generating the fused base feature $\mathbf{X}^{B} \in \mathbb{R}^{T\times 2D}$, expressed as:
            \begin{equation}\label{eq1}
                \mathbf{X}^{B}=f_{base}([\mathbf{X}^{R};\mathbf{X}^{O}];\phi_{base})\in\mathbb{R}^{T \times 2D},
            \end{equation}
            where $f_{base}$ is mainly implemented with a series of temporal convolutions with ReLU activation. Additionally, an actionness attention weight $\textbf{a} \in \mathbb{R}^{T\times 1} $ is generated to differentiate between the foreground and the background regions:
            \begin{equation}\label{eq2}
                \textbf{a} = \frac{\mathcal{A}(\mathbf{X}^{R},\mathbf{X}^{O})+\mathcal{A}(\mathbf{X}^{O},\mathbf{X}^{R})}{2} \in \mathbb{R}^{T\times 1},
            \end{equation}
            where $\mathcal{A}(\cdot)$ is an attention branch consisting of several temporal convolutional layers. Following the MIL framework, the base feature $\mathbf{X}^{B}$ is projected into the classification head $f_{cls}$ to generate the base class activation sequence (CAS):
            \begin{equation}\label{eq3}
                \textbf{S}^{base}=f_{cls}(\mathbf{X}^{B};\phi_{cls}) \in{} \mathbb{R}^{T \times (C+1)},
            \end{equation}
            where $C$ is the number of action classes. To handle background regions in untrimmed videos, we add an auxiliary class to model the background. We then aggregate snippet-level activation scores to obtain video-level class prediction $ \textbf{p}^{base}=\mathcal{K}(\textbf{S}^{base})\in \mathbb{R}^{C+1}$, where $\mathcal{K}(\cdot)$ represents the top-k average pooling along the temporal axis. After obtaining the video-level category prediction, we construct a loss function $\mathcal{L}_{base}$ using cross-entropy loss as follows:
            \begin{equation}\label{eq4}
                \mathcal{L}_{base}=-\sum_{c=1}^{C+1}\textbf{y}^{base}\:\text{log}\:(\textbf{p}_{c}^{base}),
            \end{equation}
            where $\textbf{y}^{base}=[y_{1},\cdots,y_{C},1]\in \mathbb{R}^{C+1}$ is the video-level label with an auxiliary background class. Background-suppressed CAS can be acquired by applying the attention weight $\text{S}_{supp}=\textbf{a} \otimes \textbf{S}^{base}$. We can also build a loss function $\mathcal{L}_{supp}$ with background-suppressed video-level class score $\textbf{p}^{supp}=\mathcal{K}(\text{S}_{supp}) \in \mathbb{R}^{C+1}$ as follows:
            \begin{equation}\label{eq5}
                \mathcal{L}_{supp}=-\sum_{c=1}^{C+1}\textbf{y}^{supp}\:\text{log}\:(\textbf{p}_{c}^{supp}),
            \end{equation}
            where $\textbf{y}^{supp}=[y_{1},\cdots,y_{C},0]\in \mathbb{R}^{C+1}$ is the video-level label without a background class. Optimizing $\mathcal{L}_{cls}=\mathcal{L}_{base}+\mathcal{L}_{supp}$ enables the model to distinguish snippets that significantly contribute to video-level action classification. 
            
            Our proposed method can be applied to any base head based on the MIL approach. We adopt $\text{CO}_{2}$-Net~\cite{hong2021cross} as the foundation to derive video-level class predictions, chosen for its simplicity and well-documented code. Additionally, we incorporate the previously introduced $\mathcal{L}_{oppo}$, $\mathcal{L}_{norm}$, and $\mathcal{L}_{guide}$ to enhance the optimization of the base head. As these losses were proposed in previous works~\cite{islam2021hybrid, min2020adversarial, paul2018w, lee2020background, lee2021weakly}, we do not claim originality for them. The overall objective of the baseline approach $\mathcal{L}_{vid}$ is defined as:
            \begin{equation}\label{eq6}
            \mathcal{L}_{vid}=\lambda_{1}\mathcal{L}_{cls} + \lambda_{2}\mathcal{L}_{oppo} + \lambda_{3}\mathcal{L}_{norm} + \lambda_{4}\mathcal{L}_{guide}.
            \end{equation}

    \subsection{Probabilistic Class Activation Sequence}\label{sec3b}
        In this section, we reformulate CAS into a probabilistic class activation sequence (P-CAS) to effectively utilize VLP knowledge~\cite{CLIP} within a probabilistic embedding space. To achieve this, we model the probabilistic distribution $p_{\mathbf{z}|\mathbf{x}}(\mathbf{z}|\theta)$ and estimate the parameters $\theta$, optimizing neural networks using human action and VLP knowledge. The full framework is depicted in Figure~\ref{fig:fig2}.  

        \subsubsection{Probabilistic Embedding}\label{sec3ba}
            Initially, we establish a probabilistic embedding space by leveraging pre-trained human action knowledge on Kinetics~\cite{i3d}. From the base feature $\mathbf{X}^{B}=\left\{\mathbf{x}_{t} \right\}_{t=1}^{T} \in \mathbb{R}^{T \times 2D}$, we formulate a snippet-level probability distribution $p(\mathbf{z}|\mathbf{x}_{t})$ as a multivariate Gaussian distribution with a mean vector and a diagonal covariance matrix to model the probabilistic embedding space:
                \begin{equation}\label{eq7}
                    p(\mathbf{z}|\mathbf{x}_{t})\approx \mathcal{N}(g_{\mu}(\mathbf{x}_{t}),\textup{diag}(g_{\Sigma}(\mathbf{x}_{t}))),
                \end{equation}
            where $g_{\mu}$ is an embedding layer that estimates the mean vector $g_{\mu}(\mathbf{x}_{t})\in\mathbb{R}^{D}$ and $g_{\Sigma}$ is an embedding layer that estimates covariance matrix $g_{\Sigma}(\mathbf{x}_{t}) \in \mathbb{R}^{D}$ of the target Gaussian. With the estimated $p(\mathbf{z}|\mathbf{x}_{t})$, we can sample $K$ random embeddings $\mathbf{z}^{(k)} \in \mathbb{R}^{D}$ that can represent the estimated distribution following~\cite{kingma2015variational}:
                \begin{equation}\label{eq8}
                    \mathbf{z}_{t}^{(k)} = g_{\mu}(\mathbf{x}_{t}) + \epsilon^{(k)}\cdot g_{\Sigma}(\mathbf{x}_{t}) \in \mathbb{R}^{D},
                \end{equation} 
            where $\epsilon^{(k)} \in \mathbb{R}^{D}$ are independently and identically sampled from a $D$-dimensional unit Gaussian. Our goal is to utilize $K$ embeddings sampled from the estimated probability distribution for each snippet to capture human actions from a more diverse range of perspectives. Additionally, for textual information, we transform action category names into pre-trained embeddings. We can achieve this by freezing the CLIP text transformer $\Psi_{\mathcal{C}}(\cdot)$ and extracting the embeddings $\mathbf{X}^{\mathcal{C}}=\left\{\mathbf{x}_{c} \right\}_{c=1}^{C+1}$:
                \begin{equation}\label{eq9}
                    \mathbf{x}_{c}=\Psi_{\mathcal{C}}([\mathbf{L}_{s};\Psi_{emb}(t_{c});\mathbf{L}_{e}])\in \mathbb{R}^{D},
                \end{equation}
            where $\mathbf{L}_{s}, \mathbf{L}_{e}$ are learnable tokens, $t_{c}$ refers to action category, and $\Psi_{emb}$ is the word embedding layer. P-CAS is then defined by determining the action confidence score along the temporal axis with the estimated probability distribution and action category representation. Specifically, to measure the confidence score between the estimated distribution and category representation, we formulate P-CAS as $\mathbf{S}_{prob}\in \mathbb{R}^{T \times (C+1)}$ via Monte-Carlo estimation:
                \begin{equation}\label{eq10}
                    s_{prob}(t,c) \approx \frac{1}{K}\sum_{k=1}^{K}\textup{sim}(\mathbf{z}_{t}^{(k)},\mathbf{x}_{c})/ \tau,
                \end{equation}
            where $\textup{sim}(\cdot)$ means cosine similarity and $\tau$ is the temperature parameter. Besides, to enhance the differentiation among action categories, we design an orthogonal loss $\mathcal{L}_{ortho}$ to ensure the uniqueness of each category representation:
                    \begin{equation}\label{eq11}
                    \mathcal{L}_{ortho}=\left\|\mathbf{X}^{\mathcal{C}} (\mathbf{X}^{\mathcal{C}})^{\rm T} - \mathbf{I} \right\|_{\rm F}^{2},
                    \end{equation}
            where $\mathbf{I}$ is the identity matrix and $\left\|\cdot \right\|_{\rm F}^{2}$ is the Frobenius norm of a matrix. 

        \subsubsection{VLP Knowledge Distillation}\label{sec3bb}
            However, during the estimation of the current probability distribution, only the textual information from VLP is utilized, overlooking the alignment between human action knowledge and the visual representation provided by VLP. Therefore, we aim to integrate VLP visual knowledge into the probability distribution estimation process. 
            
            Estimating the entire distribution can be challenging due to the deterministic pre-training of CLIP. However, large-scale pre-trained representations can offer a generalized point approximation (\textit{e.g.}, mean vector) for the desired distribution. To achieve this, our probabilistic embedding utilizes the CLIP's deterministic representations as estimates for the mean, $g_{\mu}(\mathbf{x}_{t})$ of the targeted distribution. To transfer VLP knowledge into the probabilistic embedding space, we first sample a set of frames $\left\{f_{t} \right\}_{t=1}^{T}$ with a fixed temporal stride from the video. Next, we freeze the CLIP Image Encoder $\Psi_{\mathcal I}(\cdot)$ and extract the embeddings $\mathbf{X}^{\mathcal I}=\left\{\Psi_{\mathcal I}(f_{t}) \right\}_{t=1}^{T} \in \mathbb{R}^{T \times D}$. For a pair of the base snippet feature and CLIP image feature $(\mathbf{x}_{t}, \mathbf{x}_{t}^{i})$, the distillation loss $\mathcal{L}_{kd}$ is defined as:
                \begin{equation}\label{eq12}
                    \mathcal{L}_{kd}=-\frac{1}{T}\sum_{t=1}^{T} \log(\frac{1}{2}(\frac{g_{\mu}(\mathbf{x}_{t})\cdot \mathbf{x}_{t}^{i}}{\left\|g_{\mu}(\mathbf{x}_{t}) \right\| \left\|\mathbf{x}_{t}^{i} \right\|}+1)).
                \end{equation}
                
            We utilize the rescaled cosine similarity between the estimated mean $g_{\mu}(\mathbf{x}_{t}^{b})$ and CLIP image representation $\mathbf{x}_{t}^{i}$ as the matching score. The objective of $\mathcal{L}_{kd}$ is to align the estimated mean $g_{\mu}(\mathbf{x}_{t}^{b})$ with the generalized fixed point $\mathbf{x}_{t}^{i}$ of CLIP embedding, thus transferring pre-trained CLIP knowledge into the desired probability distribution. 
            
    \subsection{Distribution Contrastive Learning}\label{sec3c}
        We formulate a probabilistic embedding space by aligning human action knowledge with VLP knowledge. However, the crucial factor of distributional similarity remains unexplored. Distributions corresponding to human actions should exhibit similarities with each other while contrasting with background distributions. To address this, we aim to enhance the completeness of the probabilistic embedding space through distribution contrastive learning based on statistical distances.

        \subsubsection{Intra-Distribution Contrastive Learning} \label{sec3ca} 
            We begin by considering contrastive learning between distributions within the video. Action distributions within a video are expected to share similarities while remaining distinct from background distributions. To achieve these objectives, we adopt the snippet mining algorithm of CoLA~\cite{zhang2021cola}, which uses attention weight $\mathbf{a} \in \mathbb{R}^{T \times 1}$ to differentiate between action and background snippets within the video. We classify easily distinguishable samples as easy actions (top-k) and easy backgrounds (bottom-k) based on the actionness score. However, boundary-adjacent snippets are less reliable due to their transitional position, making detection hard.
            To mine hard action and background snippets, we threshold the attention weight (1 indicates action, 0 indicates background):
                \begin{equation}\label{eq13}
                     \mathbf{b}^{(t)} = \begin{cases}
                        1 & \text{if} \quad \mathbf{a}^{(t)} > \theta_{b} \\
                        0 & \text{otherwise}
                        \end{cases},
                \end{equation}
            where $\theta_{b}$ is the threshold value. 
            
            We utilize the same strategy, performing cascaded dilation or erosion operations to identify challenging samples  (hard to differentiate) at the action/background boundaries.
                \begin{align}\label{eq14_15}
                     \mathcal{R}_{inner}=(\mathbf{b};m)^{-}-(\mathbf{b};\mathcal{M})^{-}
                     \\
                     \mathcal{R}_{outer}=(\mathbf{b};\mathcal{M})^{+}-(\mathbf{b};m)^{+},
                \end{align}
            where $(\cdot)^{-}$ and $(\cdot)^{+}$ represent the binary erosion and dilation operations with the smaller mask $m$ and the larger mask $\mathcal{M}$. Following earlier work~\cite{zhang2021cola}, we consider inner regions as hard action snippet sets and outer regions as hard background snippet sets. Here, we define hard actions as having a positive relation $\mathcal{P}_{act}$ with easy actions (top-k) and hard backgrounds as having a positive relation $\mathcal{P}_{bkg}$ with easy backgrounds (bottom-k). After snippet mining, we utilize KL divergence as a statistical metric to measure the similarity of snippet distributions. The KL divergence between multivariate Gaussian is defined as:
                \begin{multline}\label{eq16}
                    \text{KL}(\mathcal{N}_{P}\parallel \mathcal{N}_{Q})=\frac{1}{2}(\textup{tr}(\Sigma_{Q}^{-1} \Sigma_{P})+\\(\mu_{Q}-\mu_{P})^{\rm T}\Sigma_{Q}^{-1} (\mu_{Q}-\mu_{P}) + \textup{ln}(\frac{\det \Sigma_{Q}}{\det \Sigma_{P}})-D).
                \end{multline}
                
            To ensure intra-class compactness of the embedding space, we propose an intra-contrastive loss $\mathcal{L}_{intra}$ to refine snippet-level distribution similarity. The intra-contrastive loss $\mathcal{L}_{intra}$ is formulated as:
                \begin{equation}\label{eq17}
                    \mathcal{L}_{intra}=\begin{cases}
                    -\log(1-p(\mathcal{N})) & \text{if} \; (\mathcal{N}_{P},\mathcal{N}_{Q}) \in \mathcal{P} \\
                    -\log(p(\mathcal{N})) & \text{otherwise} 
                    \end{cases},
                \end{equation}
            where $\mathcal{P}$ represents positive sets for intra-distribution contrastive learning and $(\mathcal{N}_p, \mathcal{N}_Q)$ represents two arbitrary Gaussians. We formulate the matching probability $p(\mathcal{N})$ as the Jensen-Shannon divergence $\text{JSD}(\mathcal{N}_{P},\mathcal{N}_{Q})$, based on KL divergence.
            
    \begin{table*}[ht!]
    \centering
    \captionsetup{justification=justified,singlelinecheck=false}
    \caption{Comparison with previous state-of-the-art methods on THUMOS14. 0.1:0.7 and 0.1:0.5 represent the average mAP under IoU thresholds of 0.1:0.7 and 0.1:0.5.}
        \begin{tabular}{c|l|l|ccccccc|cc}
        \toprule
        \multicolumn{1}{c|}{\multirow{2}{*}[-.3em]{Supervision}}&\multicolumn{1}{c|}{\multirow{2}{*}[-.3em]{Method}} & \multicolumn{1}{c|}{\multirow{2}{*}[-.3em]{Venue}}  & \multicolumn{7}{c}{mAP@IoU (\%)}       & \multicolumn{2}{|c}{AVG} \\ \cmidrule(){4-12}
            &  &  & 0.1 & 0.2 & 0.3 & 0.4 & 0.5 & 0.6 & 0.7 & 0.1:0.7 & 0.1:0.5 \\ \midrule
            \multirow{3}{*}{\begin{tabular}[c]{@{}c@{}}Fully\\ supervised\end{tabular}}& TAL-Net~\cite{chao2018rethinking} & CVPR 2018  & 59.8 & 57.1 & 53.2 & 48.5 & 42.8 & 33.8 & 20.8 &	45.1 & 52.3 \\	
            & P-GCN~\cite{zeng2019graph} & CVPR 2019 & 69.5 & 67.8 & 63.6 & 57.8 & 49.1 & - & - & - & 61.6	\\	
            & BUMR~\cite{zhao2020bottom} & ECCV 2020 & - & - & 53.9 & 50.7 & 45.4 & 38.0 & 28.5 &	- & -	\\	\midrule
            \multirow{13}{*}{\begin{tabular}[c]{@{}c@{}}Weakly\\ supervised\end{tabular}}& CoLA~\cite{zhang2021cola} & CVPR 2021 & 66.2 & 59.5 & 51.5 &	41.9 & 32.2 & 22.0 & 13.1 &	40.9 & 50.3	\\	
            & $\text{CO}_{2}$-Net~\cite{hong2021cross}& MM 2021 & 70.1& 63.6 & 54.5 & 45.7& 38.3 &	26.4 & 13.4 & 44.6 &54.4	\\ \cmidrule(){2-12}
            & Xu \textit{et al}.~\cite{xu2023bilateral} & TPAMI 2023 & 73.1 &	66.9 & 58.3 & 48.8 & 36.5 &	24.4 & 13.4 & 45.9 &56.7	\\	
            & Li \textit{et al}.~\cite{li2023boosting} & CVPR 2023 &	71.1 & 65.0 & 56.2 & 47.8 &	39.3 & 27.5 & 15.2 & 46.0 &55.9 	\\
            & Wang \textit{et al}.~\cite{wang2023two}&CVPR 2023&	73.0&	68.2&	60.0&	47.9&	37.1&	24.4&	12.7&	46.2&57.2 \\
            & P-MIL~\cite{ren2023proposal}&	CVPR 2023& 71.8&	67.5&	58.9&	49.0&	40.0&	27.1&	15.1&	47.0&57.4	\\	
            & AHLM~\cite{wang2023weakly}&ICCV 2023&	\textbf{75.1}&	68.9&	60.2&	48.9&	38.3&	26.8&	14.7&	47.2&	58.3	\\
            & DDG-Net~\cite{tang2023ddg}&ICCV 2023& 72.5&	67.7&	58.2&	49.0&	41.4&	27.6&	14.8&	47.3&57.8\\
            & STCL-Net~\cite{fu2023semantic}&TPAMI 2023&	72.7&	67.1&	58.2&	49.7&	41.8&	28.7&	16.0&	47.7&	57.9 \\
            & GauFuse~\cite{zhou2023improving}& CVPR 2023 &	74.0&	\underline{69.4}&	60.7&	51.8&	\underline{42.7}&	26.2&	13.1&	48.3&59.7 \\ 
            & Chen \textit{et al}. ~\cite{ju2023distilling}&CVPR 2023&	73.5&	68.8&	\textbf{61.5}&	\textbf{53.8}&	42.0&	\underline{29.4}&	\underline{16.8}&	\underline{49.4}&\underline{60.0}\\ 
            & ISSF~\cite{yun2024weakly}&AAAI 2024&	72.4&	66.9&	58.4&	49.7&	41.8&	25.5&	12.8&	46.8&	57.8	\\ \cmidrule(){2-12}
            & \textbf{PVLR} (Ours) & \multicolumn{1}{l|}{MM 2024} & \underline{74.9}&	\textbf{69.9}&	\underline{61.4}&	\underline{53.1}&	\textbf{45.1}&	\textbf{30.5}&	\textbf{17.1}&	\textbf{50.3}&	\textbf{60.9}\\
        \bottomrule
        \end{tabular}
    \label{tab:thumos}
    \end{table*}

        \subsubsection{Inter-Distribution Contrastive Learning} \label{sec3cb}  
            We further introduce inter-distribution contrastive learning utilizing action category labels to ensure inter-class separability. Here, we represent the whole video distribution $p(\mathbf{z}|V)$ as a Gaussian mixture model (GMM) to measure video-level similarity,
                \begin{equation}\label{eq18}
                    p(\mathbf{z}|V)\approx \sum_{t=1}^{T} \textbf{a}_{t} \cdot \mathcal{N}(g_{\mu}(\mathbf{x}_{t}),\textup{diag}(g_{\Sigma}(\mathbf{x}_{t}))).
                \end{equation}
            To estimate $p(\mathbf{z}|V)$, we use the attention weight $\mathbf{a} \in \mathbb{R}^{T}$ as a mixing coefficient to appropriately combine distributions based on the actionness score. Given video-level category labels, we formulate a self-similarity map $ \textbf{H}\in \mathbb{R}^{N \times N}$ (1 for the same class, 0 for different) to characterize relationships between videos. Similar to intra-contrastive learning, we compute the matching probabilities $p(\mathcal{N})$ between $N$ videos within a mini-batch across mixture models, and enhance inter-video representation by comparing them with the self-similarity map. Finally, the inter-contrastive loss $\mathcal{L}_{inter}$ is formulated as:
                \begin{equation}\label{eq19}
                    \mathcal{L}_{inter}=-\frac{1}{N^{2}}\sum_{i=1}^{N}\sum_{j=1}^{N}\mathcal{L}_\text{BCE}(\textbf{H}(i,j), p(\mathcal{N})), 
                \end{equation}
            where $\mathcal{L}_{\text{BCE}}$ is a binary cross entroy loss. Beyond aligning with VLP knowledge and the probabilistic embedding space, our proposed contrastive learning framework enforces constraints on the probabilistic representation to ensure both intra-compactness and inter-separability.

    \subsection{Total Objectives} \label{sec4a}   
        Considering all the previously mentioned objectives, the total objective $\mathcal{L}_{total}$ of the entire framework is defined as:
            \begin{equation}\label{eq20}
                \mathcal{L}_{total} = \mathcal{L}_{vid} + \alpha \mathcal{L}_{kd} + \beta \mathcal{L}_{ortho} + \gamma (\mathcal{L}_{intra}+\mathcal{L}_{inter}),
            \end{equation}
        where $\alpha, \beta$, and $\gamma$ are hyper-parameters to balance these loss terms. We kept the parameters related to $\mathcal{L}_{vid}$ unchanged and performed a grid search only on $\alpha, \beta$, and $\gamma$. During testing, we use the same strategy as previous research~\cite{hong2021cross, li2023boosting} to extract proposal candidates for an input video. Lastly, we apply soft non-maximum suppression to eliminate overlapping proposals.

\section{EXPERIMENTS}
    \subsection{Experimental Settings}
        We conduct experiments on two popular WTAL benchmarks: THUMOS14~\cite{idrees2017thumos} and ActivityNet v1.3~\cite{caba2015activitynet}. THUMOS14 is a widely used benchmark for the WTAL problem, containing 200 validation videos and 213 test videos across 20 sports categories. Following previous works~\cite{hong2021cross, zhang2021cola, rizve2023pivotal}, we use the 200 validation videos for training our framework and the 213 test videos for evaluation. THUMOS14 is the most challenging dataset in WTAL due to motion blur, significant intra-class variations, and extremely short action instances. ActivityNet v1.3 has 10,024 training videos, 4,926 validation videos, and 5,044 testing videos across 200 action categories. Since the testing set annotations are not released, we train on the training set and test on the validation set. Challenges in ActivityNet typically involve the large number of action categories. Following standard evaluation metrics, we evaluate our method using mean Average Precision (mAP) under different Intersection over Union (IoU) thresholds on the temporal axis.

    \subsection{Implementation Details}
        To conduct experiments on the THUMOS14 and ActivityNet v1.3, we first divide each video into non-overlapping segments consisting of 16 frames. Subsequently, we extract the 1024-dimensional RGB and optical flow features using the I3D network~\cite{i3d} pre-trained on the Kinetics400~\cite{i3d} dataset. We apply the TV-L1 algorithm to extract the optical flow features. The fixed number of segments $T$, is set to 320 for THUMOS14 and 60 for ActivityNet v1.3. We adopt ResNet-50 as the backbone network for the CLIP image encoder $\Psi_{\mathcal I}$. It is worth noting that the I3D network and the CLIP encoders are not fine-tuned during training. For CLIP image features, we divide the video as described above, and the middle frame of each snippet is fed into $\Psi_{\mathcal I}$. For the probabilistic adapter, $g_{\mu}$ indicates a single linear layer, while $g_{\Sigma}$ is a separate network with a linear layer followed by the ReLU function, to ensure the $\Sigma$ remains positive definite. Similar to previous work ~\cite{ju2022prompting}, we prepend and append 4 prompt vectors to word embedding $\Psi_{emb}(t_{c})$, which is initialized with $\mathcal{N}(0, 0.01)$. In P-CAS, we randomly initialize the learnable embedding to model the background class, which is hard to characterize. Our experiments are conducted on an NVIDIA Tesla V100 GPU.
        
    \subsection{Comparison With State-Of-The-Art Methods}
        In this section, we compare our proposed PVLR with previous state-of-the-art methods. For THUMOS14, it is evident that the proposed PVLR outperforms all previous state-of-the-art methods, as shown in Table \ref{tab:thumos}. Notably, in WTAL scenarios where performance under high IoU (0.5-0.7) is crucial, our method surpasses all existing methodologies. In direct comparison to prior studies~\cite{li2023boosting, ju2023distilling} that also incorporate textual information, our approach demonstrates superior performance, with a margin ranging from 0.9\% to 4.3\% in average mAP (0.1:0.7). Additionally, our approach either outperforms or reaches similar performance levels to recent fully supervised methods. In Table \ref{tab:anet13}, results for the larger dataset ActivityNet v1.3 are presented. Similarly, our proposed PVLR shows superior performance compared to existing state-of-the-art methods under weakly supervised settings.

        \begin{table}[t!]
        \centering
        \captionsetup{justification=justified,singlelinecheck=false}
        \caption{Results on ActivityNet v1.3. 0.5:0.95 indicates the average mAP at IoU thresholds of 0.5:0.95.}
            \resizebox{0.95\columnwidth}{!}{
            \begin{tabular}{l|l|ccc|c}
                \toprule
                \multicolumn{1}{c|}{\multirow{2}{*}[-.3em]{Method}} & \multicolumn{1}{c|}{\multirow{2}{*}[-.3em]{Venue}}  & \multicolumn{3}{c}{mAP@IoU (\%)}       & \multicolumn{1}{|c}{AVG} \\ \cmidrule(){3-6}
                    &  & 0.5 & 0.75 & 0.95 & 0.5:0.95 \\ \midrule
                    DCC~\cite{li2022exploring}&	CVPR 2022 &	38.8&	24.2&	5.7&	24.3\\
                    RSKP~\cite{huang2022weakly}&	CVPR 2022 &	40.6&	24.6&	5.9&	25.0\\
                    ASM-Loc~\cite{he2022asm}&	CVPR 2022 &	41.0&	24.9&	6.2&	25.1\\ \midrule
                    STCL-Net~\cite{fu2023semantic}&	TPAMI 2023 &	40.6&	24.0&	6.0&	24.7\\
                    Zhang \textit{et al}.~\cite{zhang2023cross}&	TCSVT 2023 &	41.6&	25.1&	6.5&	25.3\\
                    LPR~\cite{hu2023learning}&	TCSVT 2023 &	41.4&	25.3&	6.2&	25.4\\
                    P-MIL~\cite{ren2023proposal}&	CVPR 2023 &	41.8&	25.4&	5.2&	25.5\\
                    AHLM~\cite{wang2023weakly}&	ICCV 2023 &	42.3&	24.8&	\textbf{6.9}&	25.9\\
                    Li \textit{et al}.~\cite{li2023boosting}&	CVPR 2023 &	41.8&	26.0&	6.0&	26.0\\
                    Wang \textit{et al}.~\cite{wang2023two} &	CVPR 2023 &	41.8&	25.7&	6.5&	26.3\\
                    Li \textit{et al}.~\cite{li2023weakly}&	TNNLS 2023 &	42.3&	26.4&	6.1&	26.4\\
                    CASE~\cite{liu2023revisiting}&	ICCV 2023 &	\underline{43.2}&	26.2&	\underline{6.7}&	26.8\\ \midrule
                    Yun \textit{et al}.~\cite{yun2024weakly}&	AAAI 2024 &	39.4&	25.8&	6.4&	25.8\\
                    SRHN~\cite{zhao2024snippets}&	TCSVT 2024 &	41.7&	26.1&	6.1&	26.2\\
                    Liu \textit{et al}.~\cite{liu2024modal}&	ICASSP 2024 &	42.8&	\underline{26.8}&	6.0&	26.4\\ \midrule
                    \textbf{PVLR} (Ours)&	\multicolumn{1}{l|}{MM 2024}&	\textbf{43.6}&	\textbf{27.4}&	6.5&	\textbf{27.4}\\
                \bottomrule
            \end{tabular}}
        \label{tab:anet13}
        \end{table}

    \subsection{Ablation Study}
        To demonstrate the effectiveness of our model components, we analyze the impact of each component on the THUMOS14 dataset in this section. In Table \ref{tab:abl_loss}, the baseline is reported based on solely $\mathcal{L}_{vid}$ without probabilistic embedding. The VLP knowledge distillation module serves as a pivotal step in our framework, marking the inception of our approach. By conducting feature alignment in a probabilistic space, PVLR introduces a fundamental basis that was previously overlooked in earlier literature. Integrating VLP knowledge results in a performance boost of 3.7\% through the implementation of a probabilistic class activation sequence (P-CAS). Additionally, refining our probabilistic embedding space with distribution contrastive learning leads to a 2.0\% improvement. Finally, introducing orthogonalization of text embeddings enhances the discriminative capacity between text category embeddings, yielding a 1.3\% gain. We ultimately demonstrate the effectiveness of our proposed module by achieving a performance improvement of 7.0\%, a level that is difficult to find in previous research.

        \begin{table}[t!]
        \centering
        \captionsetup{justification=justified,singlelinecheck=false}
        \caption{Component-wise ablation study on THUMOS14. }
            \resizebox{0.95\columnwidth}{!}{
            \begin{tabular}{l|ccc|c}
                \toprule
                     \multicolumn{1}{c|}{\multirow{2}{*}[-.3em]{Method}} &  \multicolumn{3}{c|}{mAP@IoU} & \multicolumn{1}{c}{AVG} \\ \cmidrule(){2-5}
                      & 0.3 & 0.5 & 0.7  & 0.3:0.7 \\ \midrule
                     Baseline & 53.0 & 36.5 & 13.6 & 34.4 \\ \midrule
                     +Distillation from VLP knowledge & 58.1 & 40.3 & 15.3 & 38.1 \\
                     +Intra-contrastive & 59.4 & 42.3 & 15.9 & 39.5 \\
                     +Inter-contrastive & 59.6 & 43.7 & 16.9 & 40.1  \\ \midrule
                     +Orthogonalization of text prompts & \textbf{61.4} & \textbf{45.1} & \textbf{17.1} & \textbf{41.4}  \\
                \bottomrule
            \end{tabular}}
            \label{tab:abl_loss}
        \end{table}

        \begin{table}[t!]
        \centering
        \captionsetup{justification=justified,singlelinecheck=false}
        \caption{Further analysis for probabilistic representation.}
            \begin{tabular}{c|ccc|c}
                 \toprule
                 \multirow{2}{*}[-.3em]{Metric} & \multicolumn{3}{c|}{mAP@IoU} & \multicolumn{1}{c}{AVG} \\ \cmidrule(){2-5}
                 & 0.3 & 0.5 & 0.7  & 0.3:0.7 \\ \midrule
                 Deterministic CAS & 57.5 & 41.0 & 15.8 & 38.3 \\ \midrule
                 Mahalanonis Distance & 60.1 & 44.0 & 17.3 & 40.7 \\
                 Bhattacharyya Distance & 60.8 & 44.0 & 17.3 & 40.9 \\
                 Kullback–Leibler Divergence & \textbf{61.4} & \textbf{45.1} & \textbf{17.1} & \textbf{41.4}  \\
                \bottomrule
            \end{tabular}
            \label{tab:abl_metric}
        \end{table}

    \subsection{Discussion}
        To provide deeper insights into the design aspect of our proposed framework, we conduct several experiments in this section. 
        
        \subsubsection{Probabilistic Representation} 
\begin{figure}[t]
            \centering
            \includegraphics[width=0.95\linewidth]{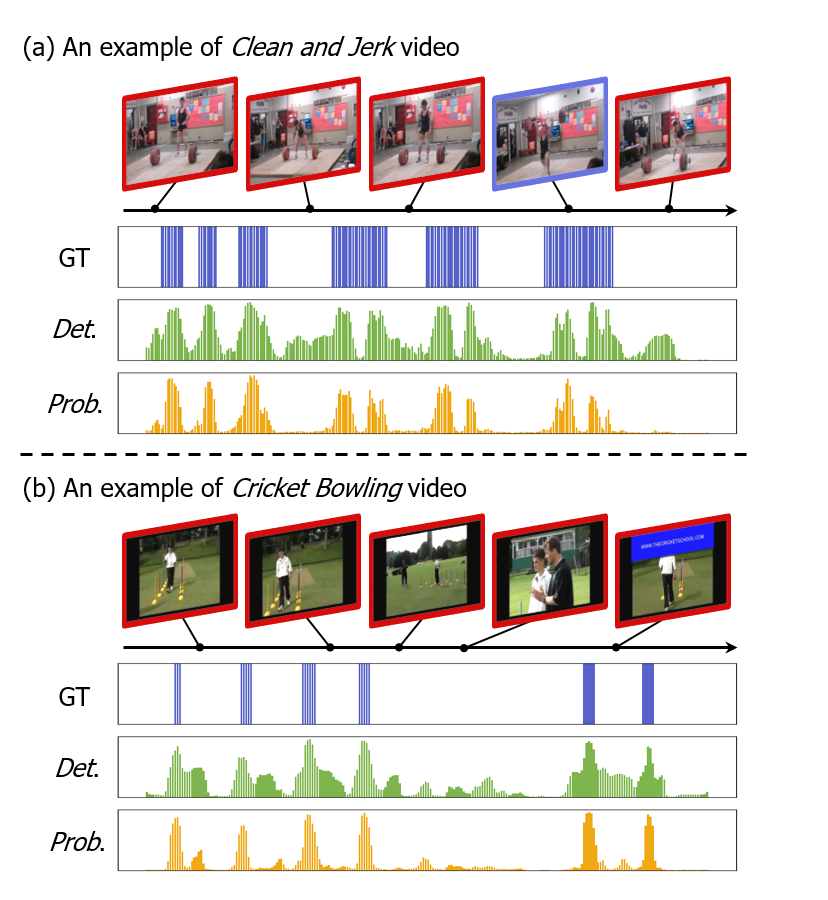}
            \captionsetup{justification=justified,singlelinecheck=false}
            \caption{Qualitative Results on THUMOS14. We compared the class activation sequence (CAS) of deterministic and probabilistic approaches. In this case, the red box is for the background, and the blue box is for the action. \vspace{-6mm}}
            \label{fig:fig3}
            \Description[Qualitative results for comparative experiments.]{This figure illustrates the prediction differences between deterministic methods and our proposed probabilistic approach based on two video examples.}
        \end{figure}

            Probabilistic representation is of great importance as the initial procedure in our framework. To validate its significance, we develop a simple baseline using deterministic representation. For the deterministic baseline, we conduct experiments utilizing one-to-one matching between human action knowledge and text embedding without a probabilistic adapter. In Table \ref{tab:abl_metric}, the first row 'Deterministic CAS' indicates the deterministic baseline. Quantitatively, there is a performance decrease of about 3.1\% compared to the proposed probabilistic approach as shown in Table \ref{tab:abl_metric}. We also compare qualitative visualization results of selected videos from the THUMOS14 dataset. Figure~\ref{fig:fig3} illustrates that the deterministic approach frequently produces predictions beyond the ground truth boundaries, struggling to capture subtle variations in human actions. In contrast, the probabilistic method effectively models temporal dynamics, focusing its predictions on the segments where real actions unfold. In Figure~\ref{fig:fig3}(a), the probabilistic approach appears to struggle with completely filling the GT segment. However, this specific area, characterized by an absence of motion change, is designated for future exploration. From Table \ref{tab:abl_metric}, besides the KL divergence, other statistical metrics are also suitable for our contrastive learning. Table \ref{tab:abl_metric} reveals that metrics capable of assessing inter-distributional similarity exhibit relatively consistent performance with minimal variation. By not relying on a specific distance metric, we can estimate a well-generalized probability distribution, leading to the successful modeling of a probabilistic embedding space. Finally, the marginal superiority of KL divergence leads to its utilization in the proposed distribution contrastive learning.

        \subsubsection{Number of\, $K$ samples} 
            To analyze the impact of the number of samples during the generation of P-CAS, we compare the performances under different values of $K$, as shown in Table \ref{tab:abl_K}. As observed, a small value for $K$ leads to suboptimal performance, resulting in a lack of representation of the estimated distribution. Here, we denote the previously described deterministic baseline as $K=0$. Since larger values of $K$ capture the entire distribution of the snippet through Monte-Carlo estimation, performance improves with an increase in $K$. Nevertheless, an increased value of $K$ results in higher computational demands. Calculating the confidence score for composing a P-CAS requires computations on the order of \,$\mathcal{O}(K)$ for each snippet and action category. Considering the computational overhead, we decide on $K=20$.
    
        \subsubsection{Computational Cost} 
            As shown in Table \ref{tab:efficiency}, we measure the model complexity in terms of additional text features (Feature), multiply-accumulative operations (MACs), the number of trainable parameters (Params), and running time (Time). Since we utilize only a lightweight probabilistic encoder (a single linear layer) with our base approach, $\text{CO}_2\text{-Net}$~\cite{hong2021cross}, it shows nearly similar complexity. Li's approach~\cite{li2023boosting}, which includes GloVe~\cite{pennington2014glove} embeddings, shows higher complexity due to the dual branch optimization. The official source code for Chen’s approach~\cite{ju2023distilling}, which also utilizes CLIP like ours, is unavailable; therefore, its complexity cannot be determined. Even though our work does not claim efficiency as its main contribution, it still demonstrates a competitive trade-off, achieving the best performance.
    
        \begin{table}[t!]
        \centering
        \captionsetup{justification=justified,singlelinecheck=false}
        \caption{Number of $K$ ablation study on THUMOS14.}
            \begin{tabular}{c|ccc|c}
                \toprule
                 \multirow{2}{*}[-.3em]{\# of samples} & \multicolumn{3}{c|}{mAP@IoU} & \multicolumn{1}{c}{AVG} \\ \cmidrule(){2-5}
                 & 0.3 & 0.5 & 0.7  & 0.3:0.7 \\ \midrule
                 Baseline & 53.0 & 36.5 & 13.6 & 34.4 \\ \midrule
                 $K=0 $ & 57.5 & 41.0 & 15.8  & 38.3 \\ 
                 $K=5$ & 59.8 & 41.4 & 15.5  & 38.9 \\
                 $K=10$ & 60.1 & 43.8 & 16.9 & 40.5 \\
                 $K=20$ & \textbf{61.4} & \textbf{45.1} & \textbf{17.1} & \textbf{41.4} \\
                \bottomrule
            \end{tabular}
            \label{tab:abl_K}
        \end{table}

        \begin{table}[ht!]
            \captionsetup{justification=justified,singlelinecheck=false}
            \caption{Computational cost comparison on THUMOS14.}
            \begin{tabular}{c|c|c|c|c|c}
            \toprule
             & Feature & Params& MACs & Time & AVG \\ \midrule
            $\text{CO}_2\text{-Net} $~\cite{hong2021cross} & - & 34.1M & 20.9G & 1.14s & 44.6 \\ \midrule
            Li~\textit{et al}.~\cite{li2023boosting} & GloVe~\cite{pennington2014glove} & 96.6M & 41.3G & 2.20s & 46.0 \\ 
            Chen~\textit{et al}.~\cite{ju2023distilling} & CLIP~\cite{CLIP} & - & - & - & 49.4 \\ \midrule
            PVLR & CLIP~\cite{CLIP} & 49.9M & 30.3G & 1.45s & \textbf{50.3} \\ 
            \bottomrule
            \end{tabular}
            \label{tab:efficiency}
        \end{table} 
    
        \begin{table}[t!]
        \centering
        \captionsetup{justification=justified,singlelinecheck=false}
        \caption{Framework generalization results on THUMOS14.}
            \begin{tabular}{c|ccc|c}
                 \toprule
                 \multirow{2}{*}[-.3em]{Method} & \multicolumn{3}{c|}{mAP@IoU} & \multicolumn{1}{c}{AVG} \\ \cmidrule(){2-5}
                 & 0.3 & 0.5 & 0.7  & 0.3:0.7 \\ \midrule
                 BaS-Net~\cite{lee2020background} & 44.6 & 26.6 & 10.0 & 27.0  \\
                 BaS-Net+Ours & 50.1 & 29.2 & 10.7 & \textbf{30.2} \\ \midrule
                 CoLA~\cite{zhang2021cola} & 51.8 & 34.0 & 12.5 & 32.9 \\
                 CoLA+Ours & 56.2 & 35.5 & 13.3 & \textbf{35.1}  \\
                \bottomrule
            \end{tabular}
            \label{tab:generalization}
        \end{table}

    \subsection{Generalization Study}
        In Table \ref{tab:generalization}, we demonstrate the generality of our contributions by integrating them into previous works in a plug-and-play manner. To achieve this, we conduct comparative experiments by replacing the base WTAL head with those from previous works~\cite{lee2020background, zhang2021cola}. The additional training modules exclusively consider the proposed probabilistic adapter for probabilistic embedding. Furthermore, we reformulate the classification objective using probabilistic class activation sequences (P-CAS). The results show that our framework enhances performance, with an average mAP increase ranging from 2\% to 3\%, indicating robust generalization across various methods and model architecture designs.

\section{Conclusion and future works}
    In this work, we present a novel framework that leverages a large-scale pre-trained vision-language model to address WTAL. Our motivation arose from the shortcomings of VLP's deterministic representation and the lack of joint alignment considerations in understanding human actions. To address these concerns, we introduce a probabilistic embedding framework aligned with human action and VLP knowledge, enhanced by distribution contrastive learning. Our method significantly outperforms previous approaches on two prominent datasets, demonstrating the effectiveness of probabilistic embedding within the VLP representation. However, the exploration of probabilistic embedding for text data represented solely by action category names remains unexplored. For future work, we will explore leveraging the recently acclaimed large-language model (LLM) to generate attributes for each action category and subsequently integrate these with probabilistic embeddings. 

\clearpage

\begin{acks}
  This work was supported by the Institute for Information and Communications Technology Planning and Evaluation (IITP) grant funded by the Korea Government (MSIT) under Grant IITP-2024-RS-2022-00156345 (ICT Challenge and Advanced Network of HRD, 50\%), Grant IITP-2024-RS-2023-00254529 (metaverse support program to nurture the best talents, 25\%) and Grant 2020-0-00011 (Video Coding for Machine, 25\%).
\end{acks}

\bibliographystyle{ACM-Reference-Format}
\bibliography{reference}


\begin{thebibliography}{61}


\ifx \showCODEN    \undefined \def \showCODEN     #1{\unskip}     \fi
\ifx \showDOI      \undefined \def \showDOI       #1{#1}\fi
\ifx \showISBNx    \undefined \def \showISBNx     #1{\unskip}     \fi
\ifx \showISBNxiii \undefined \def \showISBNxiii  #1{\unskip}     \fi
\ifx \showISSN     \undefined \def \showISSN      #1{\unskip}     \fi
\ifx \showLCCN     \undefined \def \showLCCN      #1{\unskip}     \fi
\ifx \shownote     \undefined \def \shownote      #1{#1}          \fi
\ifx \showarticletitle \undefined \def \showarticletitle #1{#1}   \fi
\ifx \showURL      \undefined \def \showURL       {\relax}        \fi
\providecommand\bibfield[2]{#2}
\providecommand\bibinfo[2]{#2}
\providecommand\natexlab[1]{#1}
\providecommand\showeprint[2][]{arXiv:#2}

\bibitem[Caba~Heilbron et~al\mbox{.}(2015)]%
        {caba2015activitynet}
\bibfield{author}{\bibinfo{person}{Fabian Caba~Heilbron}, \bibinfo{person}{Victor Escorcia}, \bibinfo{person}{Bernard Ghanem}, {and} \bibinfo{person}{Juan Carlos~Niebles}.} \bibinfo{year}{2015}\natexlab{}.
\newblock \showarticletitle{Activitynet: A large-scale video benchmark for human activity understanding}. In \bibinfo{booktitle}{\emph{Proceedings of the ieee conference on computer vision and pattern recognition}}.
\newblock


\bibitem[Carreira and Zisserman(2017)]%
        {i3d}
\bibfield{author}{\bibinfo{person}{Joao Carreira} {and} \bibinfo{person}{Andrew Zisserman}.} \bibinfo{year}{2017}\natexlab{}.
\newblock \showarticletitle{Quo vadis, action recognition? a new model and the kinetics dataset}. In \bibinfo{booktitle}{\emph{Proceedings of the IEEE/CVF Conference on Computer Vision and Pattern Recognition}}.
\newblock


\bibitem[Chao et~al\mbox{.}(2018)]%
        {chao2018rethinking}
\bibfield{author}{\bibinfo{person}{Yu-Wei Chao}, \bibinfo{person}{Sudheendra Vijayanarasimhan}, \bibinfo{person}{Bryan Seybold}, \bibinfo{person}{David~A Ross}, \bibinfo{person}{Jia Deng}, {and} \bibinfo{person}{Rahul Sukthankar}.} \bibinfo{year}{2018}\natexlab{}.
\newblock \showarticletitle{Rethinking the faster r-cnn architecture for temporal action localization}. In \bibinfo{booktitle}{\emph{Proceedings of the IEEE/CVF Conference on Computer Vision and Pattern Recognition}}.
\newblock


\bibitem[Cheng and Bertasius(2022)]%
        {cheng2022tallformer}
\bibfield{author}{\bibinfo{person}{Feng Cheng} {and} \bibinfo{person}{Gedas Bertasius}.} \bibinfo{year}{2022}\natexlab{}.
\newblock \showarticletitle{Tallformer: Temporal action localization with a long-memory transformer}. In \bibinfo{booktitle}{\emph{Proceedings of the European Conference on Computer Vision}}.
\newblock


\bibitem[Chun et~al\mbox{.}(2021)]%
        {chun2021probabilistic}
\bibfield{author}{\bibinfo{person}{Sanghyuk Chun}, \bibinfo{person}{Seong~Joon Oh}, \bibinfo{person}{Rafael~Sampaio De~Rezende}, \bibinfo{person}{Yannis Kalantidis}, {and} \bibinfo{person}{Diane Larlus}.} \bibinfo{year}{2021}\natexlab{}.
\newblock \showarticletitle{Probabilistic embeddings for cross-modal retrieval}. In \bibinfo{booktitle}{\emph{Proceedings of the IEEE/CVF Conference on Computer Vision and Pattern Recognition}}.
\newblock


\bibitem[Conde and Turgutlu(2021)]%
        {conde2021clip}
\bibfield{author}{\bibinfo{person}{Marcos~V Conde} {and} \bibinfo{person}{Kerem Turgutlu}.} \bibinfo{year}{2021}\natexlab{}.
\newblock \showarticletitle{CLIP-Art: Contrastive pre-training for fine-grained art classification}. In \bibinfo{booktitle}{\emph{Proceedings of the IEEE/CVF Conference on Computer Vision and Pattern Recognition}}.
\newblock


\bibitem[Crowson et~al\mbox{.}(2022)]%
        {crowson2022vqgan}
\bibfield{author}{\bibinfo{person}{Katherine Crowson}, \bibinfo{person}{Stella Biderman}, \bibinfo{person}{Daniel Kornis}, \bibinfo{person}{Dashiell Stander}, \bibinfo{person}{Eric Hallahan}, \bibinfo{person}{Louis Castricato}, {and} \bibinfo{person}{Edward Raff}.} \bibinfo{year}{2022}\natexlab{}.
\newblock \showarticletitle{Vqgan-clip: Open domain image generation and editing with natural language guidance}. In \bibinfo{booktitle}{\emph{Proceedings of the European Conference on Computer Vision}}.
\newblock


\bibitem[Fu et~al\mbox{.}(2023)]%
        {fu2023semantic}
\bibfield{author}{\bibinfo{person}{Jie Fu}, \bibinfo{person}{Junyu Gao}, {and} \bibinfo{person}{Changsheng Xu}.} \bibinfo{year}{2023}\natexlab{}.
\newblock \showarticletitle{Semantic and Temporal Contextual Correlation Learning for Weakly-Supervised Temporal Action Localization}.
\newblock \bibinfo{journal}{\emph{IEEE Transactions on Pattern Analysis and Machine Intelligence}} (\bibinfo{year}{2023}).
\newblock


\bibitem[He et~al\mbox{.}(2022)]%
        {he2022asm}
\bibfield{author}{\bibinfo{person}{Bo He}, \bibinfo{person}{Xitong Yang}, \bibinfo{person}{Le Kang}, \bibinfo{person}{Zhiyu Cheng}, \bibinfo{person}{Xin Zhou}, {and} \bibinfo{person}{Abhinav Shrivastava}.} \bibinfo{year}{2022}\natexlab{}.
\newblock \showarticletitle{Asm-loc: Action-aware segment modeling for weakly-supervised temporal action localization}. In \bibinfo{booktitle}{\emph{Proceedings of the IEEE/CVF Conference on Computer Vision and Pattern Recognition}}.
\newblock


\bibitem[Hong et~al\mbox{.}(2021)]%
        {hong2021cross}
\bibfield{author}{\bibinfo{person}{Fa-Ting Hong}, \bibinfo{person}{Jia-Chang Feng}, \bibinfo{person}{Dan Xu}, \bibinfo{person}{Ying Shan}, {and} \bibinfo{person}{Wei-Shi Zheng}.} \bibinfo{year}{2021}\natexlab{}.
\newblock \showarticletitle{Cross-modal consensus network for weakly supervised temporal action localization}. In \bibinfo{booktitle}{\emph{Proceedings of the 29th ACM International Conference on Multimedia}}.
\newblock


\bibitem[Hu et~al\mbox{.}(2023)]%
        {hu2023learning}
\bibfield{author}{\bibinfo{person}{Yufan Hu}, \bibinfo{person}{Jie Fu}, \bibinfo{person}{Mengyuan Chen}, \bibinfo{person}{Junyu Gao}, \bibinfo{person}{Jianfeng Dong}, \bibinfo{person}{Bin Fan}, {and} \bibinfo{person}{Hongmin Liu}.} \bibinfo{year}{2023}\natexlab{}.
\newblock \showarticletitle{Learning Proposal-aware Re-ranking for Weakly-supervised Temporal Action Localization}.
\newblock \bibinfo{journal}{\emph{IEEE Transactions on Circuits and Systems for Video Technology}} (\bibinfo{year}{2023}).
\newblock


\bibitem[Huang et~al\mbox{.}(2022)]%
        {huang2022weakly}
\bibfield{author}{\bibinfo{person}{Linjiang Huang}, \bibinfo{person}{Liang Wang}, {and} \bibinfo{person}{Hongsheng Li}.} \bibinfo{year}{2022}\natexlab{}.
\newblock \showarticletitle{Weakly supervised temporal action localization via representative snippet knowledge propagation}. In \bibinfo{booktitle}{\emph{Proceedings of the IEEE/CVF Conference on Computer Vision and Pattern Recognition}}.
\newblock


\bibitem[Idrees et~al\mbox{.}(2017)]%
        {idrees2017thumos}
\bibfield{author}{\bibinfo{person}{Haroon Idrees}, \bibinfo{person}{Amir~R Zamir}, \bibinfo{person}{Yu-Gang Jiang}, \bibinfo{person}{Alex Gorban}, \bibinfo{person}{Ivan Laptev}, \bibinfo{person}{Rahul Sukthankar}, {and} \bibinfo{person}{Mubarak Shah}.} \bibinfo{year}{2017}\natexlab{}.
\newblock \showarticletitle{The thumos challenge on action recognition for videos “in the wild”}.
\newblock \bibinfo{journal}{\emph{Computer Vision and Image Understanding}} (\bibinfo{year}{2017}).
\newblock


\bibitem[Islam et~al\mbox{.}(2021)]%
        {islam2021hybrid}
\bibfield{author}{\bibinfo{person}{Ashraful Islam}, \bibinfo{person}{Chengjiang Long}, {and} \bibinfo{person}{Richard Radke}.} \bibinfo{year}{2021}\natexlab{}.
\newblock \showarticletitle{A hybrid attention mechanism for weakly-supervised temporal action localization}. In \bibinfo{booktitle}{\emph{Proceedings of the AAAI Conference on Artificial Intelligence}}.
\newblock


\bibitem[Jo et~al\mbox{.}(2023)]%
        {jo2023vvs}
\bibfield{author}{\bibinfo{person}{Won Jo}, \bibinfo{person}{Geuntaek Lim}, \bibinfo{person}{Gwangjin Lee}, \bibinfo{person}{Hyunwoo Kim}, \bibinfo{person}{Byungsoo Ko}, {and} \bibinfo{person}{Yukyung Choi}.} \bibinfo{year}{2023}\natexlab{}.
\newblock \showarticletitle{VVS: Video-to-Video Retrieval with Irrelevant Frame Suppression}.
\newblock  (\bibinfo{year}{2023}).
\newblock


\bibitem[Ju et~al\mbox{.}(2022)]%
        {ju2022prompting}
\bibfield{author}{\bibinfo{person}{Chen Ju}, \bibinfo{person}{Tengda Han}, \bibinfo{person}{Kunhao Zheng}, \bibinfo{person}{Ya Zhang}, {and} \bibinfo{person}{Weidi Xie}.} \bibinfo{year}{2022}\natexlab{}.
\newblock \showarticletitle{Prompting visual-language models for efficient video understanding}. In \bibinfo{booktitle}{\emph{European Conference on Computer Vision}}.
\newblock


\bibitem[Ju et~al\mbox{.}(2023)]%
        {ju2023distilling}
\bibfield{author}{\bibinfo{person}{Chen Ju}, \bibinfo{person}{Kunhao Zheng}, \bibinfo{person}{Jinxiang Liu}, \bibinfo{person}{Peisen Zhao}, \bibinfo{person}{Ya Zhang}, \bibinfo{person}{Jianlong Chang}, \bibinfo{person}{Qi Tian}, {and} \bibinfo{person}{Yanfeng Wang}.} \bibinfo{year}{2023}\natexlab{}.
\newblock \showarticletitle{Distilling vision-language pre-training to collaborate with weakly-supervised temporal action localization}. In \bibinfo{booktitle}{\emph{Proceedings of the IEEE/CVF Conference on Computer Vision and Pattern Recognition}}.
\newblock


\bibitem[Kim and Cho(2022)]%
        {kim2022background}
\bibfield{author}{\bibinfo{person}{Jinah Kim} {and} \bibinfo{person}{Jungchan Cho}.} \bibinfo{year}{2022}\natexlab{}.
\newblock \showarticletitle{Background-aware robust context learning for weakly-supervised temporal action localization}.
\newblock \bibinfo{journal}{\emph{IEEE Access}} (\bibinfo{year}{2022}).
\newblock


\bibitem[Kingma et~al\mbox{.}(2015)]%
        {kingma2015variational}
\bibfield{author}{\bibinfo{person}{Durk~P Kingma}, \bibinfo{person}{Tim Salimans}, {and} \bibinfo{person}{Max Welling}.} \bibinfo{year}{2015}\natexlab{}.
\newblock \showarticletitle{Variational dropout and the local reparameterization trick}.
\newblock \bibinfo{journal}{\emph{Advances in Neural Information Processing Systems}} (\bibinfo{year}{2015}).
\newblock


\bibitem[Kwon et~al\mbox{.}(2023)]%
        {kwon2023probabilistic}
\bibfield{author}{\bibinfo{person}{Hyeongjun Kwon}, \bibinfo{person}{Taeyong Song}, \bibinfo{person}{Somi Jeong}, \bibinfo{person}{Jin Kim}, \bibinfo{person}{Jinhyun Jang}, {and} \bibinfo{person}{Kwanghoon Sohn}.} \bibinfo{year}{2023}\natexlab{}.
\newblock \showarticletitle{Probabilistic Prompt Learning for Dense Prediction}. In \bibinfo{booktitle}{\emph{Proceedings of the IEEE/CVF Conference on Computer Vision and Pattern Recognition}}.
\newblock


\bibitem[Lee et~al\mbox{.}(2020)]%
        {lee2020background}
\bibfield{author}{\bibinfo{person}{Pilhyeon Lee}, \bibinfo{person}{Youngjung Uh}, {and} \bibinfo{person}{Hyeran Byun}.} \bibinfo{year}{2020}\natexlab{}.
\newblock \showarticletitle{Background suppression network for weakly-supervised temporal action localization}. In \bibinfo{booktitle}{\emph{Proceedings of the AAAI Conference on Artificial Intelligence}}.
\newblock


\bibitem[Lee et~al\mbox{.}(2021)]%
        {lee2021weakly}
\bibfield{author}{\bibinfo{person}{Pilhyeon Lee}, \bibinfo{person}{Jinglu Wang}, \bibinfo{person}{Yan Lu}, {and} \bibinfo{person}{Hyeran Byun}.} \bibinfo{year}{2021}\natexlab{}.
\newblock \showarticletitle{Weakly-supervised temporal action localization by uncertainty modeling}. In \bibinfo{booktitle}{\emph{Proceedings of the AAAI Conference on Artificial Intelligence}}.
\newblock


\bibitem[Li et~al\mbox{.}(2023a)]%
        {li2023weakly}
\bibfield{author}{\bibinfo{person}{Guozhang Li}, \bibinfo{person}{De Cheng}, \bibinfo{person}{Xinpeng Ding}, \bibinfo{person}{Nannan Wang}, \bibinfo{person}{Jie Li}, {and} \bibinfo{person}{Xinbo Gao}.} \bibinfo{year}{2023}\natexlab{a}.
\newblock \showarticletitle{Weakly Supervised Temporal Action Localization With Bidirectional Semantic Consistency Constraint}.
\newblock \bibinfo{journal}{\emph{IEEE Transactions on Neural Networks and Learning Systems}} (\bibinfo{year}{2023}).
\newblock


\bibitem[Li et~al\mbox{.}(2023b)]%
        {li2023boosting}
\bibfield{author}{\bibinfo{person}{Guozhang Li}, \bibinfo{person}{De Cheng}, \bibinfo{person}{Xinpeng Ding}, \bibinfo{person}{Nannan Wang}, \bibinfo{person}{Xiaoyu Wang}, {and} \bibinfo{person}{Xinbo Gao}.} \bibinfo{year}{2023}\natexlab{b}.
\newblock \showarticletitle{Boosting Weakly-Supervised Temporal Action Localization with Text Information}. In \bibinfo{booktitle}{\emph{Proceedings of the IEEE/CVF Conference on Computer Vision and Pattern Recognition}}.
\newblock


\bibitem[Li et~al\mbox{.}(2022)]%
        {li2022exploring}
\bibfield{author}{\bibinfo{person}{Jingjing Li}, \bibinfo{person}{Tianyu Yang}, \bibinfo{person}{Wei Ji}, \bibinfo{person}{Jue Wang}, {and} \bibinfo{person}{Li Cheng}.} \bibinfo{year}{2022}\natexlab{}.
\newblock \showarticletitle{Exploring denoised cross-video contrast for weakly-supervised temporal action localization}. In \bibinfo{booktitle}{\emph{Proceedings of the IEEE/CVF Conference on Computer Vision and Pattern Recognition}}.
\newblock


\bibitem[Lin et~al\mbox{.}(2019)]%
        {lin2019bmn}
\bibfield{author}{\bibinfo{person}{Tianwei Lin}, \bibinfo{person}{Xiao Liu}, \bibinfo{person}{Xin Li}, \bibinfo{person}{Errui Ding}, {and} \bibinfo{person}{Shilei Wen}.} \bibinfo{year}{2019}\natexlab{}.
\newblock \showarticletitle{Bmn: Boundary-matching network for temporal action proposal generation}. In \bibinfo{booktitle}{\emph{Proceedings of the IEEE/CVF International Conference on Computer Vision}}.
\newblock


\bibitem[Liu et~al\mbox{.}(2019)]%
        {liu2019novel}
\bibfield{author}{\bibinfo{person}{Jichao Liu}, \bibinfo{person}{Chuanxu Wang}, {and} \bibinfo{person}{Yun Liu}.} \bibinfo{year}{2019}\natexlab{}.
\newblock \showarticletitle{A novel method for temporal action localization and recognition in untrimmed video based on time series segmentation}.
\newblock \bibinfo{journal}{\emph{IEEE Access}} (\bibinfo{year}{2019}).
\newblock


\bibitem[Liu et~al\mbox{.}(2024)]%
        {liu2024modal}
\bibfield{author}{\bibinfo{person}{Peng Liu}, \bibinfo{person}{Chuanxu Wang}, {and} \bibinfo{person}{Min Zhao}.} \bibinfo{year}{2024}\natexlab{}.
\newblock \showarticletitle{Modal Consensus and Contextual Separation for Weakly Supervised Temporal Action Localization}. In \bibinfo{booktitle}{\emph{IEEE International Conference on Acoustics, Speech and Signal Processing}}.
\newblock


\bibitem[Liu et~al\mbox{.}(2023)]%
        {liu2023revisiting}
\bibfield{author}{\bibinfo{person}{Qinying Liu}, \bibinfo{person}{Zilei Wang}, \bibinfo{person}{Shenghai Rong}, \bibinfo{person}{Junjie Li}, {and} \bibinfo{person}{Yixin Zhang}.} \bibinfo{year}{2023}\natexlab{}.
\newblock \showarticletitle{Revisiting Foreground and Background Separation in Weakly-supervised Temporal Action Localization: A Clustering-based Approach}. In \bibinfo{booktitle}{\emph{Proceedings of the IEEE/CVF International Conference on Computer Vision}}.
\newblock


\bibitem[Luo et~al\mbox{.}(2022)]%
        {luo2022clip4clip}
\bibfield{author}{\bibinfo{person}{Huaishao Luo}, \bibinfo{person}{Lei Ji}, \bibinfo{person}{Ming Zhong}, \bibinfo{person}{Yang Chen}, \bibinfo{person}{Wen Lei}, \bibinfo{person}{Nan Duan}, {and} \bibinfo{person}{Tianrui Li}.} \bibinfo{year}{2022}\natexlab{}.
\newblock \showarticletitle{Clip4clip: An empirical study of clip for end to end video clip retrieval and captioning}.
\newblock \bibinfo{journal}{\emph{Neurocomputing}} (\bibinfo{year}{2022}).
\newblock


\bibitem[Ma et~al\mbox{.}(2022)]%
        {ma2022x}
\bibfield{author}{\bibinfo{person}{Yiwei Ma}, \bibinfo{person}{Guohai Xu}, \bibinfo{person}{Xiaoshuai Sun}, \bibinfo{person}{Ming Yan}, \bibinfo{person}{Ji Zhang}, {and} \bibinfo{person}{Rongrong Ji}.} \bibinfo{year}{2022}\natexlab{}.
\newblock \showarticletitle{X-clip: End-to-end multi-grained contrastive learning for video-text retrieval}. In \bibinfo{booktitle}{\emph{Proceedings of the 30th ACM International Conference on Multimedia}}.
\newblock


\bibitem[Min and Corso(2020)]%
        {min2020adversarial}
\bibfield{author}{\bibinfo{person}{Kyle Min} {and} \bibinfo{person}{Jason~J Corso}.} \bibinfo{year}{2020}\natexlab{}.
\newblock \showarticletitle{Adversarial background-aware loss for weakly-supervised temporal activity localization}. In \bibinfo{booktitle}{\emph{Proceedings of the European Conference on Computer Vision}}.
\newblock


\bibitem[Nguyen et~al\mbox{.}(2018)]%
        {nguyen2018weakly}
\bibfield{author}{\bibinfo{person}{Phuc Nguyen}, \bibinfo{person}{Ting Liu}, \bibinfo{person}{Gautam Prasad}, {and} \bibinfo{person}{Bohyung Han}.} \bibinfo{year}{2018}\natexlab{}.
\newblock \showarticletitle{Weakly supervised action localization by sparse temporal pooling network}. In \bibinfo{booktitle}{\emph{Proceedings of the IEEE/CVF Conference on Computer Vision and Pattern Recognition}}.
\newblock


\bibitem[Novack et~al\mbox{.}(2023)]%
        {novack2023chils}
\bibfield{author}{\bibinfo{person}{Zachary Novack}, \bibinfo{person}{Julian McAuley}, \bibinfo{person}{Zachary~Chase Lipton}, {and} \bibinfo{person}{Saurabh Garg}.} \bibinfo{year}{2023}\natexlab{}.
\newblock \showarticletitle{Chils: Zero-shot image classification with hierarchical label sets}. In \bibinfo{booktitle}{\emph{International Conference on Machine Learning}}.
\newblock


\bibitem[Parelli et~al\mbox{.}(2023)]%
        {parelli2023clip}
\bibfield{author}{\bibinfo{person}{Maria Parelli}, \bibinfo{person}{Alexandros Delitzas}, \bibinfo{person}{Nikolas Hars}, \bibinfo{person}{Georgios Vlassis}, \bibinfo{person}{Sotirios Anagnostidis}, \bibinfo{person}{Gregor Bachmann}, {and} \bibinfo{person}{Thomas Hofmann}.} \bibinfo{year}{2023}\natexlab{}.
\newblock \showarticletitle{CLIP-Guided Vision-Language Pre-training for Question Answering in 3D Scenes}. In \bibinfo{booktitle}{\emph{Proceedings of the IEEE/CVF Conference on Computer Vision and Pattern Recognition}}.
\newblock


\bibitem[Park et~al\mbox{.}(2022)]%
        {park2022probabilistic}
\bibfield{author}{\bibinfo{person}{Jungin Park}, \bibinfo{person}{Jiyoung Lee}, \bibinfo{person}{Ig-Jae Kim}, {and} \bibinfo{person}{Kwanghoon Sohn}.} \bibinfo{year}{2022}\natexlab{}.
\newblock \showarticletitle{Probabilistic representations for video contrastive learning}. In \bibinfo{booktitle}{\emph{Proceedings of the IEEE/CVF Conference on Computer Vision and Pattern Recognition}}.
\newblock


\bibitem[Paul et~al\mbox{.}(2018)]%
        {paul2018w}
\bibfield{author}{\bibinfo{person}{Sujoy Paul}, \bibinfo{person}{Sourya Roy}, {and} \bibinfo{person}{Amit~K Roy-Chowdhury}.} \bibinfo{year}{2018}\natexlab{}.
\newblock \showarticletitle{W-talc: Weakly-supervised temporal activity localization and classification}. In \bibinfo{booktitle}{\emph{Proceedings of the European Conference on Computer Vision}}.
\newblock


\bibitem[Pennington et~al\mbox{.}(2014)]%
        {pennington2014glove}
\bibfield{author}{\bibinfo{person}{Jeffrey Pennington}, \bibinfo{person}{Richard Socher}, {and} \bibinfo{person}{Christopher~D Manning}.} \bibinfo{year}{2014}\natexlab{}.
\newblock \showarticletitle{Glove: Global vectors for word representation}. In \bibinfo{booktitle}{\emph{Proceedings of the 2014 conference on Empirical Methods in Natural Language Processing}}.
\newblock


\bibitem[Radford et~al\mbox{.}(2021)]%
        {CLIP}
\bibfield{author}{\bibinfo{person}{Alec Radford}, \bibinfo{person}{Jong~Wook Kim}, \bibinfo{person}{Chris Hallacy}, \bibinfo{person}{Aditya Ramesh}, \bibinfo{person}{Gabriel Goh}, \bibinfo{person}{Sandhini Agarwal}, \bibinfo{person}{Girish Sastry}, \bibinfo{person}{Amanda Askell}, \bibinfo{person}{Pamela Mishkin}, \bibinfo{person}{Jack Clark}, {et~al\mbox{.}}} \bibinfo{year}{2021}\natexlab{}.
\newblock \showarticletitle{Learning transferable visual models from natural language supervision}. In \bibinfo{booktitle}{\emph{International Conference on Machine Learning}}.
\newblock


\bibitem[Rao et~al\mbox{.}(2022)]%
        {rao2022denseclip}
\bibfield{author}{\bibinfo{person}{Yongming Rao}, \bibinfo{person}{Wenliang Zhao}, \bibinfo{person}{Guangyi Chen}, \bibinfo{person}{Yansong Tang}, \bibinfo{person}{Zheng Zhu}, \bibinfo{person}{Guan Huang}, \bibinfo{person}{Jie Zhou}, {and} \bibinfo{person}{Jiwen Lu}.} \bibinfo{year}{2022}\natexlab{}.
\newblock \showarticletitle{Denseclip: Language-guided dense prediction with context-aware prompting}. In \bibinfo{booktitle}{\emph{Proceedings of the IEEE/CVF Conference on Computer Vision and Pattern Recognition}}.
\newblock


\bibitem[Ren et~al\mbox{.}(2023)]%
        {ren2023proposal}
\bibfield{author}{\bibinfo{person}{Huan Ren}, \bibinfo{person}{Wenfei Yang}, \bibinfo{person}{Tianzhu Zhang}, {and} \bibinfo{person}{Yongdong Zhang}.} \bibinfo{year}{2023}\natexlab{}.
\newblock \showarticletitle{Proposal-Based Multiple Instance Learning for Weakly-Supervised Temporal Action Localization}. In \bibinfo{booktitle}{\emph{Proceedings of the IEEE/CVF Conference on Computer Vision and Pattern Recognition}}.
\newblock


\bibitem[Rizve et~al\mbox{.}(2023)]%
        {rizve2023pivotal}
\bibfield{author}{\bibinfo{person}{Mamshad~Nayeem Rizve}, \bibinfo{person}{Gaurav Mittal}, \bibinfo{person}{Ye Yu}, \bibinfo{person}{Matthew Hall}, \bibinfo{person}{Sandra Sajeev}, \bibinfo{person}{Mubarak Shah}, {and} \bibinfo{person}{Mei Chen}.} \bibinfo{year}{2023}\natexlab{}.
\newblock \showarticletitle{PivoTAL: Prior-Driven Supervision for Weakly-Supervised Temporal Action Localization}. In \bibinfo{booktitle}{\emph{Proceedings of the IEEE/CVF Conference on Computer Vision and Pattern Recognition}}.
\newblock


\bibitem[Saharia et~al\mbox{.}(2022)]%
        {saharia2022photorealistic}
\bibfield{author}{\bibinfo{person}{Chitwan Saharia}, \bibinfo{person}{William Chan}, \bibinfo{person}{Saurabh Saxena}, \bibinfo{person}{Lala Li}, \bibinfo{person}{Jay Whang}, \bibinfo{person}{Emily~L Denton}, \bibinfo{person}{Kamyar Ghasemipour}, \bibinfo{person}{Raphael Gontijo~Lopes}, \bibinfo{person}{Burcu Karagol~Ayan}, \bibinfo{person}{Tim Salimans}, {et~al\mbox{.}}} \bibinfo{year}{2022}\natexlab{}.
\newblock \showarticletitle{Photorealistic text-to-image diffusion models with deep language understanding}.
\newblock \bibinfo{journal}{\emph{Advances in Neural Information Processing Systems}} (\bibinfo{year}{2022}).
\newblock


\bibitem[Shen et~al\mbox{.}(2020)]%
        {shen2020weakly}
\bibfield{author}{\bibinfo{person}{Zhengyang Shen}, \bibinfo{person}{Feng Wang}, {and} \bibinfo{person}{Jin Dai}.} \bibinfo{year}{2020}\natexlab{}.
\newblock \showarticletitle{Weakly supervised temporal action localization by multi-stage fusion network}.
\newblock \bibinfo{journal}{\emph{IEEE Access}} (\bibinfo{year}{2020}).
\newblock


\bibitem[Shou et~al\mbox{.}(2018)]%
        {shou2018autoloc}
\bibfield{author}{\bibinfo{person}{Zheng Shou}, \bibinfo{person}{Hang Gao}, \bibinfo{person}{Lei Zhang}, \bibinfo{person}{Kazuyuki Miyazawa}, {and} \bibinfo{person}{Shih-Fu Chang}.} \bibinfo{year}{2018}\natexlab{}.
\newblock \showarticletitle{Autoloc: Weakly-supervised temporal action localization in untrimmed videos}. In \bibinfo{booktitle}{\emph{Proceedings of the European Conference on Computer Vision}}.
\newblock


\bibitem[Tang et~al\mbox{.}(2023)]%
        {tang2023ddg}
\bibfield{author}{\bibinfo{person}{Xiaojun Tang}, \bibinfo{person}{Junsong Fan}, \bibinfo{person}{Chuanchen Luo}, \bibinfo{person}{Zhaoxiang Zhang}, \bibinfo{person}{Man Zhang}, {and} \bibinfo{person}{Zongyuan Yang}.} \bibinfo{year}{2023}\natexlab{}.
\newblock \showarticletitle{DDG-Net: Discriminability-Driven Graph Network for Weakly-supervised Temporal Action Localization}. In \bibinfo{booktitle}{\emph{Proceedings of the IEEE/CVF International Conference on Computer Vision}}.
\newblock


\bibitem[Tirupattur et~al\mbox{.}(2021)]%
        {tirupattur2021modeling}
\bibfield{author}{\bibinfo{person}{Praveen Tirupattur}, \bibinfo{person}{Kevin Duarte}, \bibinfo{person}{Yogesh~S Rawat}, {and} \bibinfo{person}{Mubarak Shah}.} \bibinfo{year}{2021}\natexlab{}.
\newblock \showarticletitle{Modeling multi-label action dependencies for temporal action localization}. In \bibinfo{booktitle}{\emph{Proceedings of the IEEE/CVF Conference on Computer Vision and Pattern Recognition}}.
\newblock


\bibitem[Upadhyay et~al\mbox{.}(2023)]%
        {upadhyay2023probvlm}
\bibfield{author}{\bibinfo{person}{Uddeshya Upadhyay}, \bibinfo{person}{Shyamgopal Karthik}, \bibinfo{person}{Massimiliano Mancini}, {and} \bibinfo{person}{Zeynep Akata}.} \bibinfo{year}{2023}\natexlab{}.
\newblock \showarticletitle{Probvlm: Probabilistic adapter for frozen vison-language models}. In \bibinfo{booktitle}{\emph{Proceedings of the IEEE/CVF International Conference on Computer Vision}}.
\newblock


\bibitem[Wang et~al\mbox{.}(2023b)]%
        {wang2023weakly}
\bibfield{author}{\bibinfo{person}{Guiqin Wang}, \bibinfo{person}{Peng Zhao}, \bibinfo{person}{Cong Zhao}, \bibinfo{person}{Shusen Yang}, \bibinfo{person}{Jie Cheng}, \bibinfo{person}{Luziwei Leng}, \bibinfo{person}{Jianxing Liao}, {and} \bibinfo{person}{Qinghai Guo}.} \bibinfo{year}{2023}\natexlab{b}.
\newblock \showarticletitle{Weakly-Supervised Action Localization by Hierarchically-structured Latent Attention Modeling}. In \bibinfo{booktitle}{\emph{Proceedings of the IEEE/CVF International Conference on Computer Vision}}.
\newblock


\bibitem[Wang et~al\mbox{.}(2023a)]%
        {wang2023two}
\bibfield{author}{\bibinfo{person}{Yu Wang}, \bibinfo{person}{Yadong Li}, {and} \bibinfo{person}{Hongbin Wang}.} \bibinfo{year}{2023}\natexlab{a}.
\newblock \showarticletitle{Two-Stream Networks for Weakly-Supervised Temporal Action Localization With Semantic-Aware Mechanisms}. In \bibinfo{booktitle}{\emph{Proceedings of the IEEE/CVF Conference on Computer Vision and Pattern Recognition}}.
\newblock


\bibitem[Xu et~al\mbox{.}(2023)]%
        {xu2023bilateral}
\bibfield{author}{\bibinfo{person}{Zhe Xu}, \bibinfo{person}{Kun Wei}, \bibinfo{person}{Erkun Yang}, \bibinfo{person}{Cheng Deng}, {and} \bibinfo{person}{Wei Liu}.} \bibinfo{year}{2023}\natexlab{}.
\newblock \showarticletitle{Bilateral Relation Distillation for Weakly Supervised Temporal Action Localization}.
\newblock \bibinfo{journal}{\emph{IEEE Transactions on Pattern Analysis and Machine Intelligence}} (\bibinfo{year}{2023}).
\newblock


\bibitem[Xue et~al\mbox{.}(2023)]%
        {xue2022clip}
\bibfield{author}{\bibinfo{person}{Hongwei Xue}, \bibinfo{person}{Yuchong Sun}, \bibinfo{person}{Bei Liu}, \bibinfo{person}{Jianlong Fu}, \bibinfo{person}{Ruihua Song}, \bibinfo{person}{Houqiang Li}, {and} \bibinfo{person}{Jiebo Luo}.} \bibinfo{year}{2023}\natexlab{}.
\newblock \showarticletitle{Clip-vip: Adapting pre-trained image-text model to video-language representation alignment}.
\newblock \bibinfo{journal}{\emph{International Conference on Learning Representations}} (\bibinfo{year}{2023}).
\newblock


\bibitem[Yun et~al\mbox{.}(2024)]%
        {yun2024weakly}
\bibfield{author}{\bibinfo{person}{Wulian Yun}, \bibinfo{person}{Mengshi Qi}, \bibinfo{person}{Chuanming Wang}, {and} \bibinfo{person}{Huadong Ma}.} \bibinfo{year}{2024}\natexlab{}.
\newblock \showarticletitle{Weakly-Supervised Temporal Action Localization by Inferring Salient Snippet-Feature}. In \bibinfo{booktitle}{\emph{Proceedings of the AAAI Conference on Artificial Intelligence}}.
\newblock


\bibitem[Zeng et~al\mbox{.}(2019)]%
        {zeng2019graph}
\bibfield{author}{\bibinfo{person}{Runhao Zeng}, \bibinfo{person}{Wenbing Huang}, \bibinfo{person}{Mingkui Tan}, \bibinfo{person}{Yu Rong}, \bibinfo{person}{Peilin Zhao}, \bibinfo{person}{Junzhou Huang}, {and} \bibinfo{person}{Chuang Gan}.} \bibinfo{year}{2019}\natexlab{}.
\newblock \showarticletitle{Graph convolutional networks for temporal action localization}. In \bibinfo{booktitle}{\emph{Proceedings of the IEEE/CVF International Conference on Computer Vision}}.
\newblock


\bibitem[Zhang et~al\mbox{.}(2021)]%
        {zhang2021cola}
\bibfield{author}{\bibinfo{person}{Can Zhang}, \bibinfo{person}{Meng Cao}, \bibinfo{person}{Dongming Yang}, \bibinfo{person}{Jie Chen}, {and} \bibinfo{person}{Yuexian Zou}.} \bibinfo{year}{2021}\natexlab{}.
\newblock \showarticletitle{Cola: Weakly-supervised temporal action localization with snippet contrastive learning}. In \bibinfo{booktitle}{\emph{Proceedings of the IEEE/CVF Conference on Computer Vision and Pattern Recognition}}.
\newblock


\bibitem[Zhang et~al\mbox{.}(2022b)]%
        {zhang2022actionformer}
\bibfield{author}{\bibinfo{person}{Chen-Lin Zhang}, \bibinfo{person}{Jianxin Wu}, {and} \bibinfo{person}{Yin Li}.} \bibinfo{year}{2022}\natexlab{b}.
\newblock \showarticletitle{Actionformer: Localizing moments of actions with transformers}. In \bibinfo{booktitle}{\emph{Proceedings of the European Conference on Computer Vision}}.
\newblock


\bibitem[Zhang et~al\mbox{.}(2022a)]%
        {zhang2022temporal}
\bibfield{author}{\bibinfo{person}{Min Zhang}, \bibinfo{person}{Haiyang Hu}, {and} \bibinfo{person}{Zhongjin Li}.} \bibinfo{year}{2022}\natexlab{a}.
\newblock \showarticletitle{Temporal action localization with coarse-to-fine network}.
\newblock \bibinfo{journal}{\emph{IEEE Access}} (\bibinfo{year}{2022}).
\newblock


\bibitem[Zhang and Zhao(2023)]%
        {zhang2023cross}
\bibfield{author}{\bibinfo{person}{Songchun Zhang} {and} \bibinfo{person}{Chunhui Zhao}.} \bibinfo{year}{2023}\natexlab{}.
\newblock \showarticletitle{Cross-Video Contextual Knowledge Exploration and Exploitation for Ambiguity Reduction in Weakly Supervised Temporal Action Localization}.
\newblock \bibinfo{journal}{\emph{IEEE Transactions on Circuits and Systems for Video Technology}} (\bibinfo{year}{2023}).
\newblock


\bibitem[Zhao et~al\mbox{.}(2020)]%
        {zhao2020bottom}
\bibfield{author}{\bibinfo{person}{Peisen Zhao}, \bibinfo{person}{Lingxi Xie}, \bibinfo{person}{Chen Ju}, \bibinfo{person}{Ya Zhang}, \bibinfo{person}{Yanfeng Wang}, {and} \bibinfo{person}{Qi Tian}.} \bibinfo{year}{2020}\natexlab{}.
\newblock \showarticletitle{Bottom-up temporal action localization with mutual regularization}. In \bibinfo{booktitle}{\emph{Proceedings of the European Conference on Computer Vision}}.
\newblock


\bibitem[Zhao et~al\mbox{.}(2024)]%
        {zhao2024snippets}
\bibfield{author}{\bibinfo{person}{Yibo Zhao}, \bibinfo{person}{Hua Zhang}, \bibinfo{person}{Zan Gao}, \bibinfo{person}{Weili Guan}, \bibinfo{person}{Meng Wang}, {and} \bibinfo{person}{Shengyong Chen}.} \bibinfo{year}{2024}\natexlab{}.
\newblock \showarticletitle{A Snippets Relation and Hard-Snippets Mask Network for Weakly-Supervised Temporal Action Localization}.
\newblock \bibinfo{journal}{\emph{IEEE Transactions on Circuits and Systems for Video Technology}} (\bibinfo{year}{2024}).
\newblock


\bibitem[Zhou et~al\mbox{.}(2023)]%
        {zhou2023improving}
\bibfield{author}{\bibinfo{person}{Jingqiu Zhou}, \bibinfo{person}{Linjiang Huang}, \bibinfo{person}{Liang Wang}, \bibinfo{person}{Si Liu}, {and} \bibinfo{person}{Hongsheng Li}.} \bibinfo{year}{2023}\natexlab{}.
\newblock \showarticletitle{Improving Weakly Supervised Temporal Action Localization by Bridging Train-Test Gap in Pseudo Labels}. In \bibinfo{booktitle}{\emph{Proceedings of the IEEE/CVF Conference on Computer Vision and Pattern Recognition}}.
\newblock


\end{thebibliography}

\end{document}